\documentclass[pdflatex,sn-mathphys-ay]{sn-jnl}

\usepackage{chngcntr}
\usepackage{graphicx}%
\usepackage{multirow}%
\usepackage{amsmath,amssymb,amsfonts}%
\usepackage{amsthm}%
\usepackage{mathrsfs}%
\usepackage[title]{appendix}%
\usepackage{xcolor}%
\usepackage{textcomp}%
\usepackage{manyfoot}%
\usepackage{booktabs}%
\usepackage{algorithm}%
\usepackage{algorithmicx}%
\usepackage{algpseudocode}%
\usepackage{listings}%
\usepackage{bbm}
\usepackage{placeins}
\usepackage{xr-hyper}
\usepackage{hyperref}
\theoremstyle{thmstyleone}%
\theoremstyle{thmstyletwo}%

\theoremstyle{thmstylethree}%

\begin{document}

\title[A Measurement Perspective on Benchmarking LLM-Generated Text Detectors using Item Response Theory ]{A Measurement Perspective on Benchmarking LLM-Generated Text Detectors using Item Response Theory}


\author[1]{\fnm{Yanbin} \sur{Zhao}}\email{zhaoyb@sspu.edu.cn}

\author[2]{\fnm{Zhipeng} \sur{Chen}}\email{chenzhipeng0104@gmail.com}

\author[3]{\fnm{Guangsheng} \sur{Bao}}\email{baoguangsheng@fyust.edu.cn}

\author[1]{\fnm{Guandong} \sur{Zhang}}\email{gdzhang@sspu.edu.cn}

\author*[1]{\fnm{Yan} \sur{Li}}\email{yanli@sspu.edu.cn}

\affil[1]{\orgdiv{School of Mathematics and Statistics}, \orgname{Shanghai Polytechnic University}, \orgaddress{\street{2360 Jinhai Road}, \city{Pudong New District}, \postcode{201209}, \state{Shanghai}, \country{China}}}

\affil[2]{\orgdiv{School of Computer and Information Engineering}, \orgname{Shanghai Polytechnic University}, \orgaddress{\street{2360 Jinhai Road}, \city{Pudong New District}, \postcode{201209}, \state{Shanghai}, \country{China}}}

\affil[3]{\orgdiv{School of Computing and Artificial Intelligence}, \orgname{Fuyao University of Science and Technology}, \orgaddress{\street{No. 104 Zhihui Avenue}, \city{Fuzhou High-tech Industrial Development Zone}, \postcode{350109}, \state{Fujian}, \country{China}}}
\abstract{Benchmarking large language model (LLM)-generated text detectors commonly relies on aggregate detector-level metrics such as accuracy, F1 score, and AUC. Although useful for summarizing overall performance, these metrics provide limited insight into how benchmark composition shapes detector comparisons or which texts distinguish detectors with different response profiles. We propose a measurement-based benchmarking framework for jointly characterizing benchmark heterogeneity and detector behavior using multidimensional item response theory (IRT). By treating benchmark texts as measurement items and detectors as respondents, the framework estimates text-level discrimination and difficulty-location parameters alongside multidimensional detector abilities. We apply the framework to a benchmark of 32K human-written and machine-generated texts across four domains, evaluated by 14 detectors. The estimated IRT item-parameter space reveals distinct but overlapping human-machine distributions, with substantial variation across domains and text lengths. The human–machine label shows the strongest multivariate association with this space, while domain and text length further contribute to benchmark heterogeneity. Detector ability profiles reveal multidimensional and domain-dependent response patterns that are not fully reflected in aggregate performance rankings. As a practical application, clustering texts in the estimated IRT space yields a compact subset of 2,088 texts (6.5\% of the full 32K benchmark) that largely preserves the comparative detector-performance patterns observed on the full benchmark. Overall, the proposed framework provides a principled and interpretable methodology for benchmarking heterogeneous detection technologies while jointly characterizing the measurement properties of the texts used for their evaluation.}




\maketitle

\section{Introduction}\label{sec1}

Large language models (LLMs), including GPT, Claude, Gemini, Llama, and DeepSeek,
have been widely adopted in writing, programming, education, information retrieval, and multimodal applications. While their ability to generate fluent and coherent text has created substantial practical value, it has also blurred the boundary between human-written and machine-authored content, raising severe concerns over content authenticity, academic misconduct, and large-scale automated disinformation \citep{wu2025,vykopal2024,hazell2023,chen2024c,chen2024l,afane2024,cotton2024}. A large body of LLM-generated text detectors has consequently been proposed \citep{hashimoto2019,gehrmann2019,ippolito2020, mitchell2023,bao2023fast,yang2023,su2023,guo2024,hans2024,bafna2024,zhou2025}.
Nevertheless, reliable benchmark evaluation of these detectors remains an open challenge: reported performance heavily depends on test corpora, and cross-paper comparison is often unreliable. Aggregate metrics such as accuracy, F1 score, and AUC treat all benchmark samples equally and provide no diagnostic information about the benchmark itself \citep{mitchell2023,bao2023fast,guo2024,dugan2024}. This limits not only the assessment of individual detectors but also the systematic benchmarking and comparison of competing detection technologies. Consequently, we cannot tell whether poor performance stems from the detector’s intrinsic weakness or from biased, low-informative test samples.

These limitations suggest that benchmarking LLM-generated text detectors should be viewed not only as a classification problem, but also as a measurement problem. This perspective follows the original motivation of item response theory (IRT), which was developed in educational and psychological measurement to account for the fact that test items are not equally difficult or equally informative, and that respondents cannot be adequately characterized by total scores alone \citep{lord1968,lord1980,embretson2013}. IRT jointly models these item- and respondent-level properties, providing a natural analogue for detector evaluation. Benchmark texts differ in domain, length, style, and generation source \citep{dugan2024,wu2024,wu2025}, while detectors based on different principles may succeed and fail on different subsets of texts. Treating benchmark texts as measurement items and detectors as respondents therefore provides a principled way to characterize heterogeneous detector–text response patterns beyond a single aggregate score.

 Prior research has demonstrated the power of IRT-based evaluation for general-purpose NLP models and generative large language models, revealing instance-level properties hidden behind aggregate performance metrics \citep{lalor2016,rodriguez2021, maia_polo2024, zhou2025lo}. Despite these successes, existing work targets model-output evaluation rather than the detectors themselves. To fill this gap, this paper adopts multidimensional IRT as a unified measurement framework for evaluating LLM-generated-text detectors. We treat benchmark texts as test items and detectors as respondents, which enables us to jointly estimate text-level difficulty-location and discrimination parameters as well as multidimensional latent abilities for detectors.

Building on this framework, we analyze a benchmark containing 32K human-written and machine-generated texts evaluated using 14 representative detectors. We examine how the IRT-derived discrimination and difficulty-location parameters characterize differences between human-written and machine-generated texts, and how these patterns vary across domains and text lengths. We further study the multidimensional detector ability profiles to identify shared and detector-specific response patterns. Finally, we investigate whether the estimated IRT representation can support compact benchmark construction by selecting a smaller subset that preserves the main comparative detector-performance patterns of the full benchmark. 

The main contributions of this paper are as follows.

First, we introduce a measurement-based benchmarking framework for LLM-generated text detection evaluation, using IRT to model the detector–text response structure. Rather than treating detection solely as a binary classification task, the framework treats benchmark texts themselves as objects of measurement, enabling joint evaluation of benchmark heterogeneity and detector behavior.

Second, we show that the IRT item parameters provide sample-level diagnostics of benchmark heterogeneity. Across 32K human-written and machine-generated texts, the IRT item-parameter space exhibits a distinct yet overlapping human-machine structure, indicating that the two classes are not completely separable. Domain and text length further modulate this structure, demonstrating that benchmark heterogeneity cannot be fully characterized by observable attributes alone.

Third, we characterize and compare detectors through multidimensional IRT ability profiles rather than a single performance ranking. The estimated abilities reveal shared and detector-specific response patterns, differences between supervised and zero-shot methods, and domain-dependent behavior, showing that heterogeneous detection approaches cannot be fully compared through aggregate metrics alone.

Fourth, we demonstrate the practical value of the IRT representation for compact benchmark construction. A subset of 2,088 samples (6.5\% of the full 32K benchmark) preserves the main comparative detector-performance patterns observed across 14 detectors, suggesting that IRT-based sample representations can support substantially more compact evaluation benchmarks.

\subsection{Related work}

\textbf{LLM detectors.}
Existing LLM-generated text detectors can be broadly divided into supervised and zero-shot methods. Supervised detectors learn decision boundaries from labeled human-written and machine-generated texts, including RoBERTa-based detectors \citep{bafna2024}, BiScope \citep{guo2024}, and AdaDetectGPT \citep{zhou2025}. Their main advantage is strong predictive performance when training and test distributions are well matched, but their generalizability may degrade when applied to unseen domains, new LLMs, or different prompting and editing settings. 
Zero-shot detectors instead rely on statistical signals derived from language models, such as Likelihood, Entropy, Rank, LogRank, or related token-level measures \citep{hashimoto2019,gehrmann2019,ippolito2020,su2023}, or use perturbation- and model-comparison-based strategies such as DetectGPT \citep{mitchell2023}, Fast-DetectGPT \citep{bao2023fast}, and Binoculars \citep{hans2024}. These methods require little or no labeled training data and are therefore more flexible in new evaluation settings, but their performance can be sensitive to the reference model, text domain, generation strategy, and text length. These complementary strengths and limitations indicate that detector performance depends not only on the method itself, but also strongly on the characteristics of the evaluation data.

\textbf{LLM-generated text detection benchmarks.}
Given this dependence on evaluation conditions, benchmark construction plays a central role in LLM-generated text detection. Existing benchmarks range from task-oriented datasets, such as HC3 \citep{guo2023} and CHEAT \citep{yu2025}, to larger multi-domain and multi-generator corpora such as Deepfake Text Detection \citep{pu2023}, DetectRL \citep{wu2024}, and MAGE \citep{li2024mage}. Task-oriented benchmarks offer greater control over prompts, tasks, and generation procedures, but their conclusions may be tied to specific data settings. Larger and more diverse benchmarks improve coverage across domains, generators, and writing conditions, yet greater size and diversity do not necessarily guarantee evaluation quality. Most existing benchmarks are still assessed primarily through aggregate detector-level metrics, which provide limited information about the measurement properties of individual texts. In particular, they do not directly indicate whether benchmark samples span a sufficiently broad range of detection difficulty or whether they effectively distinguish detectors with different capabilities. This motivates a measurement framework that evaluates not only detector performance, but also the benchmark texts themselves.

\textbf{Evaluation and benchmarking beyond aggregate metrics.} Conventional benchmarking typically aggregates sample-level metrics such as accuracy, F1 score, and AUC. Although effective for summarizing overall performance, these metrics provide limited information about benchmark sample properties and heterogeneous detector-response patterns. Multidimensional IRT (MIRT) \citep{reckase1983} extends unidimensional IRT \citep{lord1968} by allowing responses to depend on multiple latent traits, making it suitable when performance cannot be explained by a single ability dimension. This is particularly relevant to LLM-generated text detection, where detectors may rely on heterogeneous signals, such as likelihood patterns, perturbation sensitivity, supervised textual features, or model-specific statistical traces. A key advantage of IRT is that it jointly characterizes respondents and items within a unified measurement framework. In the standard two-parameter logistic (2PL) IRT model, item difficulty reflects how difficult an item is, while the discrimination parameter measures how effectively it distinguishes respondents with different ability levels.

IRT has been widely applied beyond traditional psychometrics. It has also been increasingly adopted in AI and NLP to analyze model performance, instance difficulty, and example-level informativeness \citep{lalor2016,martinez2019,meir2024}. More recently, it has been applied to LLM evaluation for efficient model assessment \citep{maia_polo2024} and model comparison \citep{zhou2025lo}. However, existing studies have mainly focused on evaluating generative models or general NLP systems, while relatively little attention has been paid to jointly characterizing detector technologies and the benchmark samples on which they are evaluated. In detector evaluation, detectors can naturally be viewed as respondents and texts as items, with each detector–text pair yielding a binary correctness response. This formulation allows MIRT to support measurement-based benchmarking by jointly characterizing text-level difficulty and discrimination while capturing multidimensional differences in detector abilities.

\section{Measurement-based benchmark framework}

\subsection{Detector–text response formulation}\label{sec2_1}

We formulate the benchmarking of LLM-generated text detectors as a detector-text response modeling problem. Our objective is to characterize both detector behavior and the benchmark texts used for their evaluation, rather than to develop a new detector. Consider a benchmark of $I$ texts evaluated by $L$ detectors. Let $Y_i$ denote the ground-truth label of text $i$, where $Y_i=1$ indicates machine-generated text and $Y_i=0$ indicates human-written text, and $D_{il}$ denote the binary prediction of detector $l$ for text $i$. 

Rather than reducing each detector to a single aggregate performance metric, we retain the sample-level correctness through $
R_{il}=\mathbb{I}_{\{D_{il}=Y_i\}},
$ where $R_{il}=1$ indicates correct classification and $R_{il}=0$ otherwise. The resulting response matrix $R=(R_{il})$ summarizes the complete detector–text correctness structure.

This formulation provides a natural measurement perspective: texts are treated as measurement items and detectors as respondents. The response matrix therefore captures both heterogeneity among texts in how they challenge and distinguish detectors and variation among detectors in their response capabilities. MIRT is then used to jointly model these two sources of heterogeneity. For detectors producing continuous scores, the calibration procedure used to obtain $D_{il}$ is described in the experimental setup.

\subsection{Multidimensional IRT model}
To model the detector–text response structure, we adopt a compensatory multidimensional 2PL-style parameterization with a vector-valued difficulty-location parameter:
\begin{eqnarray}
p_{il}\triangleq \mathbb{P}(R_{il}=1\mid\theta_l,\alpha_i,\beta_i)
=\frac{1}{1+\exp\{-\alpha_i^{T}(\theta_l-\beta_i)\}},
\end{eqnarray}
where $\theta_l\in\mathbb{R}^{d}$ denotes the multidimensional latent ability vector of detector $l$, $\alpha_i\in\mathbb{R}^{d}$ denotes the discrimination vector of text $i$, and $\beta_i\in\mathbb{R}^{d}$ denotes its difficulty-location vector.

Under this parameterization, $\beta_i$ is not uniquely identifiable from the likelihood, since replacing $\beta_i$ by $\beta_i+v$ leaves the response probability unchanged whenever $\alpha_i^T v=0$. We retain the vector form because it represents difficulty relative to different latent ability dimensions, which would be lost under a scalar difficulty parameterization. Accordingly, the components of $\beta_i$ are interpreted as relative difficulty locations within the fitted latent coordinate system rather than absolute difficulty measures. Specifically, $\beta_{ik}$ represents the relative difficulty location of text $i$ along dimension $k$, while $\alpha_i^T\beta_i$ provides an effective scalar difficulty summary along its discrimination direction.

The discrimination vector $\alpha_i$ characterizes how strongly the correct-classification probability depends on different ability dimensions. Its direction indicates the relevant latent ability dimensions, while its magnitude reflects the overall discrimination strength. The ability vector $\theta_l$ captures multidimensional detector behavior. Thus, MIRT provides a common latent representation of text-level difficulty and discrimination and detector-level response heterogeneity.

\subsection{Bayesian estimation and model selection}
Following \citet{maia_polo2024}, we estimate the MIRT parameters using a hierarchical Bayesian procedure. We assume that $\theta_l\sim N(\mu_{\theta} \mathbbm{1}_d, 1/u_{\theta}I_d), \alpha_i\sim N (\mu_{\alpha}\mathbbm{1}_d,1/u_{\alpha}I_d)$, and $\beta_i\sim N(\mu_{\beta}\mathbbm{1}_d, 1/u_{\beta}I_d),$ with hyperpriors $\mu_{\theta}\sim N(0,10), u_{\theta}\sim \Gamma(1,1), \mu_{\alpha}\sim N(0,10), u_{\alpha}\sim \Gamma(1,1), \mu_{\beta}\sim N(0,10),$ and $u_{\beta}\sim \Gamma(1,1).$ The model is fitted by variational inference, and the posterior means are used as point estimates of $\theta_l$, $\alpha_i$, and $\beta_i$. Because the latent dimensions are identifiable only up to rotation, a post-estimation rotation is applied to improve interpretability.

We select the latent dimension from $d\in\{2,3,5,10,15\}$ using five-fold cross-validation, with binary cross-entropy between the observed correctness responses and predicted probabilities as the validation criterion. We choose $d=5$ based on validation performance while retaining a parsimonious and interpretable latent representation.

\subsection{IRT-guided compact benchmark construction}
The estimated IRT item parameters also provide a measurement-based representation for compact benchmark construction. For each text $i$, we define 
$E_i=(\hat{\alpha}_i,\hat{\beta}_i)$,
where $\hat{\alpha}_i$ and $\hat{\beta}_i$ are the estimated discrimination and difficulty-location vectors.

We apply $K$-means clustering to $\{E_i\}_{i=1}^{I}$ and select samples closest to the cluster centers as representative samples. This procedure aims to preserve the main structure of the fitted IRT space while reducing benchmark size. It is therefore used as an IRT-guided representative subset selection strategy rather than a method for selecting only the most informative or difficult individual texts.

\section{Experiments and Results}
In this section, we describe the benchmark construction and detector setup, followed by analyses of detector performance, IRT item parameters, detector latent abilities, and compact benchmark construction.

\subsection{Benchmark construction and detector setup}\label{sec3_1}

\textbf{Data source.} To evaluate the proposed measurement-based benchmarking framework, we constructed a multi-domain benchmark covering diverse writing styles and content types. Specifically, we collected 16,000 human-written texts from four domains \cite{wu2024}: ArXiv abstracts published between 2002 and 2017 (\url{https://www.kaggle.com/datasets/spsayakpaul/arxiv-paper-abstracts/data}), Writing Prompt creative stories \citep{fan2018}, XSum news articles \citep{narayan2018}, and Yelp Reviews \citep{zhang2015}. For each domain, 4,000 texts of no more than 512 words were randomly selected, representing scientific writing, creative storytelling, news summarization, and customer reviews, respectively. Because these texts predate the emergence of ChatGPT, contamination by LLM-generated content is expected to be limited. The selected texts were subsequently used as source materials for generating matched machine-generated counterparts under controlled prompting conditions.

\textbf{Machine-generated text models.} Using the human-written texts as source material, we generated machine-generated samples using GPT-5, Claude Sonnet 4.5, Qwen3-32B, and Llama 3.3 70B. For each domain, the 4,000 human-written texts were randomly divided into four subsets of 1,000 texts, each assigned to one LLM, yielding 1,000 generated texts per model per domain and 16,000 machine-generated texts in total. Following \citet{bao2023fast}, the first 30 words of each human-written text were used as the prompt prefix, preserving topical and contextual alignment while allowing the LLM to generate the continuation independently.

\textbf{Detector evaluation and response matrix construction.}
We evaluated the resulting 32,000 texts using 14 LLM-generated text detectors, including 11 zero-shot or training-free methods: DetectGPT \citep{mitchell2023}, Likelihood \citep{gehrmann2019}, Entropy \citep{gehrmann2019}, Lastde \citep{xu2025}, Rank \citep{gehrmann2019}, NPR \citep{su2023}, LogRank \citep{gehrmann2019}, Fast-DetectGPT \citep{bao2023fast}, Binoculars \citep{hans2024}, DNA-GPT \citep{yang2023}, and LRR \citep{su2023}; and three supervised RoBERTa-based classifiers: RoBERTa-Base \citep{liu2019}, RoBERTa-Large \citep{liu2019}, and RoBERTa-ChatGPT \citep{guo2023}. GPT-Neo-2.7B was used as the scoring model for white-box zero-shot detectors following \citet{bao2023fast}, while the supervised detectors used publicly available pretrained checkpoints without further retraining.

For detectors producing continuous scores, binary predictions were obtained using detector-specific thresholds selected by maximizing the Youden index \citep{youden1950}:
$$
\tau_l^*=\arg\max_{\tau}\left\{TPR_l(\tau)-FPR_l(\tau)\right\},
$$
where
$$
TPR_l(\tau)=\frac{TP_l(\tau)}{TP_l(\tau)+FN_l(\tau)}, \qquad
FPR_l(\tau)=\frac{FP_l(\tau)}{FP_l(\tau)+TN_l(\tau)}
$$
denote the true positive rate and false positive rate, respectively. 

This benchmark-level calibration maps heterogeneous detector scores onto a common binary response scale for subsequent IRT analysis. The resulting predictions were converted into correctness responses as defined in Section \ref{sec2_1}, yielding the detector–text response matrix used for subsequent MIRT analysis.

\subsection{Benchmark composition}
The final benchmark contains 32,000 texts, evenly divided between human-written and machine-generated samples. For each of the four domains, ArXiv, Writing Prompt, XSum, and Yelp Reviews, we include 4,000 human-written texts and 4,000 machine-generated texts, resulting in a balanced domain and class structure. Table \ref{tab1:samplesize} summarizes the original data source sizes and the composition of the evaluation benchmark across the four domains.

\begin{figure}
    \centering  \includegraphics[width=0.8\linewidth]{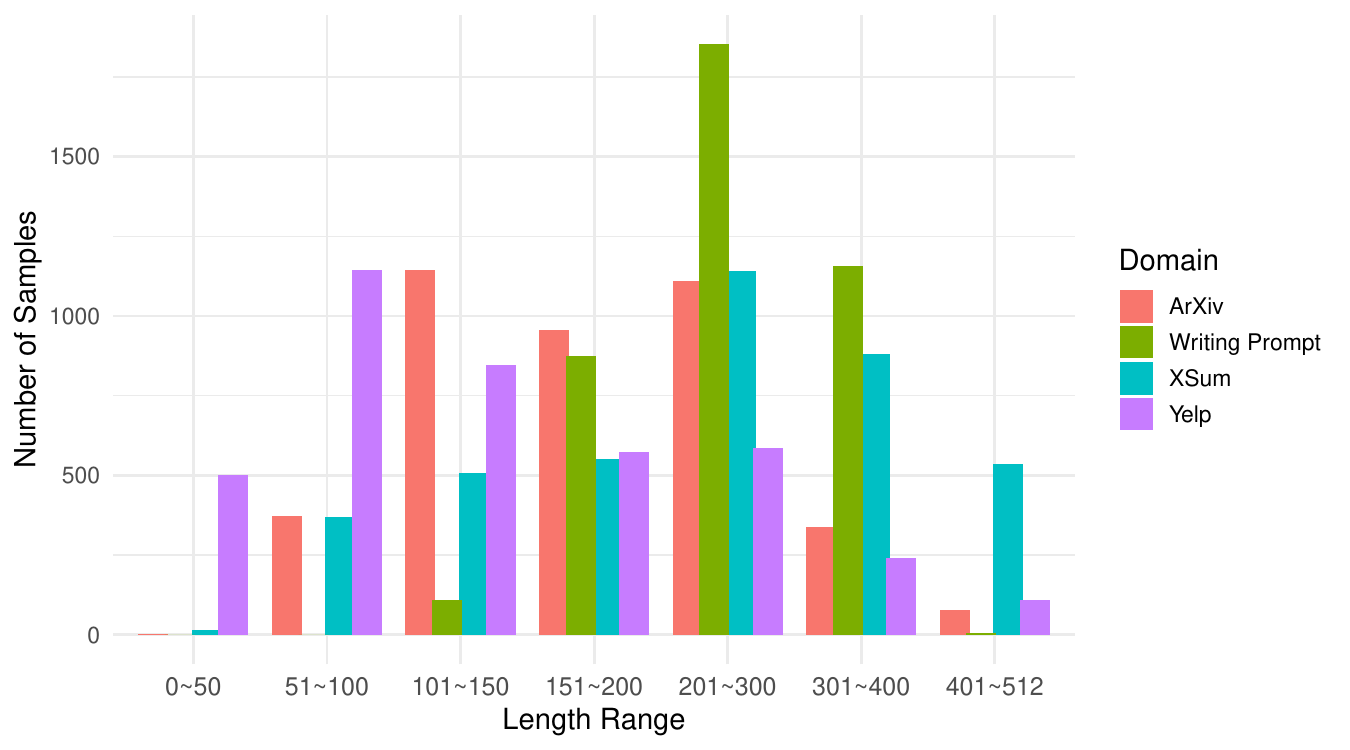}
    \caption{Sample distributions across text length ranges in the four domains.}
    \label{fig:multidomain}
\end{figure}

Despite this balance, text length varies substantially across domains (Figure \ref{fig:multidomain}), reflecting differences in writing styles and content structures. Because detector behavior may depend on text length, we examine both domain-specific and length-stratified IRT patterns.

\begin{table}[t]
\centering
    \caption{Overview of data sources and benchmark composition across four domains.}\label{tab1:samplesize}
\begin{tabular}{@{}lcccc@{}}
    \toprule
       Domains  & Original source size & Human-written & Machine-generated & Total selected\\
       \midrule
       ArXiv  & 18,891  & 4,000 &4,000&8,000\\
       Writing Prompt  & 87,485 & 4,000&4,000&8,000\\
       XSum & 2,040,445& 4,000&4,000&8,000\\
       Yelp &650,000& 4,000&4,000&8,000\\
       Total&-&16,000&16,000&32,000\\
       \bottomrule       
    \end{tabular}
\end{table}

\begin{table*}[ht]
\centering
  \caption{Overall detection performance of different detectors on the sampled benchmark, measured by AUC, F1 score, and accuracy. Numbers in parentheses indicate the rank of each detector under the corresponding metric.}  \label{tab:overall}
    \begin{tabular}{@{}lccc@{}}
    \toprule
    \textbf{Detector} & \textbf{AUC}   & \textbf{F1}    & \textbf{ACC} \\
    \midrule
    DetectGPT & 0.742 (1) & 0.626 (14) & 0.625 (3) \\
    Likelihood & 0.642 (2) & 0.672 (7) & 0.627 (2) \\
    Lastde & 0.628 (3) & 0.688 (2) & 0.617 (4) \\
    LogRank & 0.621 (4) & 0.703 (1) & 0.656 (1) \\
    Rank  & 0.614 (5) & 0.680 (4) & 0.595 (8) \\
    NPR   & 0.604 (6) & 0.653 (10) & 0.576 (10) \\
    Entropy & 0.603 (7) & 0.635 (13) & 0.606 (6) \\
    Fast-DetectGPT & 0.600 (8) & 0.686 (3) & 0.608 (5) \\
    Binoculars   & 0.590 (9) & 0.679 (5) & 0.602 (7) \\
    DNA-GPT & 0.587 (10) & 0.638 (11) & 0.591 (9) \\
    LRR   & 0.538 (11) & 0.672 (6) & 0.573 (11) \\
    RoBERTa-ChatGPT & 0.507 (12) & 0.637 (12) & 0.522 (12) \\
    RoBERTa-Base & 0.507 (13) & 0.661 (9) & 0.515 (13) \\
    RoBERTa-Large & 0.421 (14) & 0.665 (8) & 0.505 (14) \\
    \bottomrule
    \end{tabular}%

\end{table*}%
\subsection{Conventional benchmarking of detector performance}

As a conventional baseline for the proposed measurement-based benchmarking framework, we first evaluate the performance of existing detectors on the sampled benchmark. Tables \ref{tab:overall} and \ref{tab:eachdomain} report overall and domain-specific detector performance in terms of AUC, F1 score, and accuracy (ACC). AUC is threshold-independent, whereas F1 score and accuracy use the detector-specific calibration described in Section \ref{sec3_1}. Overall, the detectors achieve only moderate performance on this benchmark. As shown in Table \ref{tab:overall}, DetectGPT achieves the highest AUC (0.742), indicating relatively stronger ranking ability in distinguishing human-written from machine-generated texts. In contrast, LogRank achieves the highest F1 score (0.703) and accuracy (0.656), suggesting a better balance between precision and recall under the selected classification threshold. In comparison, the supervised RoBERTa-based detectors (RoBERTa-ChatGPT, RoBERTa-Base, and RoBERTa-Large) generally show weaker overall performance, suggesting limited robustness in heterogeneous and mixed-domain settings.

The results in Table \ref{tab:eachdomain} further show that the detectors' performance varies substantially across domains. On ArXiv, Rank and NPR detectors perform particularly well, achieving the highest AUC values of 0.876 and 0.863, respectively, as well as strong F1 scores of 0.796 and 0.786, but perform substantially worse in other domains. On Writing Prompt, DetectGPT and Fast-DetectGPT show stronger performance than most of the other competing methods. On XSum and Yelp Reviews, Likelihood and Entropy perform better in terms of AUC, whereas RoBERTa-Large and RoBERTa-Base achieve the highest F1 scores, with values of 0.666 and 0.664 in XSum, and 0.669 and 0.659 in Yelp Reviews, respectively. These findings suggest that different detectors rely on different textual signals and therefore exhibit domain-specific strengths and weaknesses. Overall, detection on the pooled benchmark remains challenging, even for the strongest methods. More importantly, detector rankings change substantially across domains and evaluation metrics, indicating that conclusions from conventional benchmarking depend on both the evaluation metric and benchmark composition. This motivates an item-level analysis of the response matrix, where benchmark samples are characterized by their difficulty and discriminative value to complement aggregate detector-level comparisons.

\begin{sidewaystable}[!htbp]
  \caption{Domain-specific detection performance of different detectors on the sampled benchmark across ArXiv, Writing Prompt, XSum, and Yelp Reviews, measured by AUC, F1 score, and accuracy. Numbers in parentheses indicate the rank of each detector under the corresponding metric within each domain.}  \label{tab:eachdomain}%

    \begin{tabular}{lcccccc}
\toprule
          & \multicolumn{3}{c}{ArXiv} & \multicolumn{3}{c}{Writing Prompt} \\
             \midrule
    \multicolumn{1}{c}{\textbf{Detector}} & \textbf{AUC} & \textbf{F1} & \textbf{ACC} & \textbf{AUC} & \textbf{F1} & \textbf{ACC} \\
\midrule
    Rank  & 0.876 (1) & 0.796 (1) & 0.799 (1) & 0.509 (10) & 0.214 (12) & 0.518 (12) \\
    NPR   & 0.863 (2) & 0.786 (2) & 0.790 (2) & 0.501 (12) & 0.094 (14) & 0.511 (13) \\
    Likelihood & 0.816 (3) & 0.755 (3) & 0.749 (3) & 0.598 (6) & 0.452 (8) & 0.587 (6) \\
    LogRank & 0.809 (4) & 0.743 (4) & 0.741 (4) & 0.557 (7) & 0.333 (11) & 0.572 (7) \\
    Entropy & 0.799 (5) & 0.734 (5) & 0.734 (5) & 0.514 (9) & 0.352 (9) & 0.519 (11) \\
    LRR   & 0.741 (6) & 0.676 (6) & 0.684 (6) & 0.440 (13) & 0.203 (13) & 0.539 (8) \\
    Lastde & 0.735 (7) & 0.665 (8) & 0.673 (7) & 0.665 (4) & 0.543 (6) & 0.641 (3) \\
    DetectGPT & 0.722 (8) & 0.672 (7) & 0.664 (8) & 0.797 (1) & 0.714 (1) & 0.725 (1) \\
    DNA-GPT & 0.687 (9) & 0.643 (11) & 0.645 (9) & 0.628 (5) & 0.485 (7) & 0.599 (5) \\
    Fast-DetectGPT & 0.681 (10) & 0.609 (12) & 0.631 (10) & 0.673 (2) & 0.547 (5) & 0.647 (2) \\
    Binoculars  & 0.661 (11) & 0.566 (13) & 0.625 (11) & 0.667 (3) & 0.564 (4) & 0.636 (4) \\
    RoBERTa-ChatGPT & 0.614 (12) & 0.046 (14) & 0.511 (12) & 0.507 (11) & 0.336 (10) & 0.527 (10) \\
    RoBERTa-Base & 0.413 (13) & 0.644 (10) & 0.481 (14) & 0.520 (8) & 0.676 (2) & 0.538 (9) \\
    RoBERTa-Large & 0.335 (14) & 0.663 (9) & 0.498 (13) & 0.423 (14) & 0.663 (3) & 0.502 (14) \\
    \bottomrule
          & \multicolumn{3}{c}{XSum} & \multicolumn{3}{c}{Yelp Reviews} \\    \midrule
    \multicolumn{1}{c}{\textbf{Detector}} & \textbf{AUC} & \textbf{F1} & \textbf{ACC} & \textbf{AUC} & \textbf{F1} & \textbf{ACC} \\
    \midrule
    Rank  & 0.556 (6) & 0.413 (10) & 0.546 (9) & 0.511 (10) & 0.143 (13) & 0.515 (13) \\
    NPR   & 0.537 (9) & 0.416 (9) & 0.533 (11) & 0.496 (11) & 0.040 (14) & 0.506 (14) \\
    Likelihood & 0.623 (2) & 0.575 (3) & 0.598 (1) & 0.588 (2) & 0.420 (6) & 0.575 (5) \\
    LogRank & 0.598 (3) & 0.437 (8) & 0.584 (3) & 0.552 (6) & 0.293 (10) & 0.559 (8) \\
    Entropy & 0.623 (1) & 0.558 (4) & 0.593 (2) & 0.605 (1) & 0.586 (4) & 0.577 (4) \\
    LRR   & 0.515 (12) & 0.296 (14) & 0.535 (10) & 0.452 (14) & 0.207 (12) & 0.536 (10) \\
    Lastde & 0.552 (8) & 0.354 (11) & 0.565 (7) & 0.576 (5) & 0.382 (7) & 0.591 (1) \\
    DetectGPT & 0.565 (5) & 0.512 (5) & 0.566 (5) & 0.583 (3) & 0.592 (3) & 0.557 (9) \\
    DNA-GPT & 0.555 (7) & 0.492 (6) & 0.549 (8) & 0.582 (4) & 0.477 (5) & 0.572 (6) \\
    Fast-DetectGPT & 0.535 (10) & 0.314 (13) & 0.569 (4) & 0.531 (7) & 0.366 (9) & 0.587 (2) \\
    Binoculars  & 0.531 (11) & 0.347 (12) & 0.566 (6) & 0.523 (8) & 0.377 (8) & 0.579 (3) \\
    RoBERTa-ChatGPT & 0.459 (13) & 0.447 (7) & 0.491 (14) & 0.488 (12) & 0.248 (11) & 0.560 (7) \\
    RoBERTa-Base & 0.586 (4) & 0.664 (2) & 0.513 (12) & 0.519 (9) & 0.659 (2) & 0.529 (11) \\
    RoBERTa-Large & 0.434 (14) & 0.666 (1) & 0.504 (13) & 0.465 (13) & 0.669 (1) & 0.517 (12) \\\bottomrule
    \end{tabular}%
\end{sidewaystable}%


\begin{table*}[!htbp]
  \caption{MANOVA results for the globally estimated IRT parameter space.}
\label{tab:manova_global_irt}
\begin{tabular}{lrrrr}
\toprule
Effect & Df & Wilks' $\Lambda$ & Approx. $F$  & $p$-value \\
\midrule
Domain & 3 & 0.88950 & 127.3 & $<0.001$ \\
Length group & 2 & 0.97000 & 49.1 &$<0.001$ \\
Group & 1 & 0.37217 & 5392.7 & $<0.001$ \\
Domain $\times$ Length group & 6 & 0.98805 & 6.4 & $<0.001$ \\
Domain $\times$ Group & 3 & 0.83174 & 202.6 & $<0.001$ \\
Length group $\times$ Group & 2 & 0.96048 & 65.1 &  $<0.001$ \\
Domain $\times$ Length group $\times$ Group & 6 & 0.98129 & 10.1  & $<0.001$ \\
\bottomrule
\end{tabular}
\end{table*}

\subsection{IRT discrimination and difficulty-location parameters}

\subsubsection{Global IRT parameter space}

We first examine the heterogeneity of the benchmark using the estimated item-level IRT parameters. The discrimination parameters $\alpha_i$ characterize how strongly each text differentiates detectors with different abilities, whereas the difficulty-location parameters $\beta_i$ characterize the relative position of each text within the fitted latent detector-ability space.

\begin{figure}
    \centering
    \includegraphics[width=0.75\linewidth]{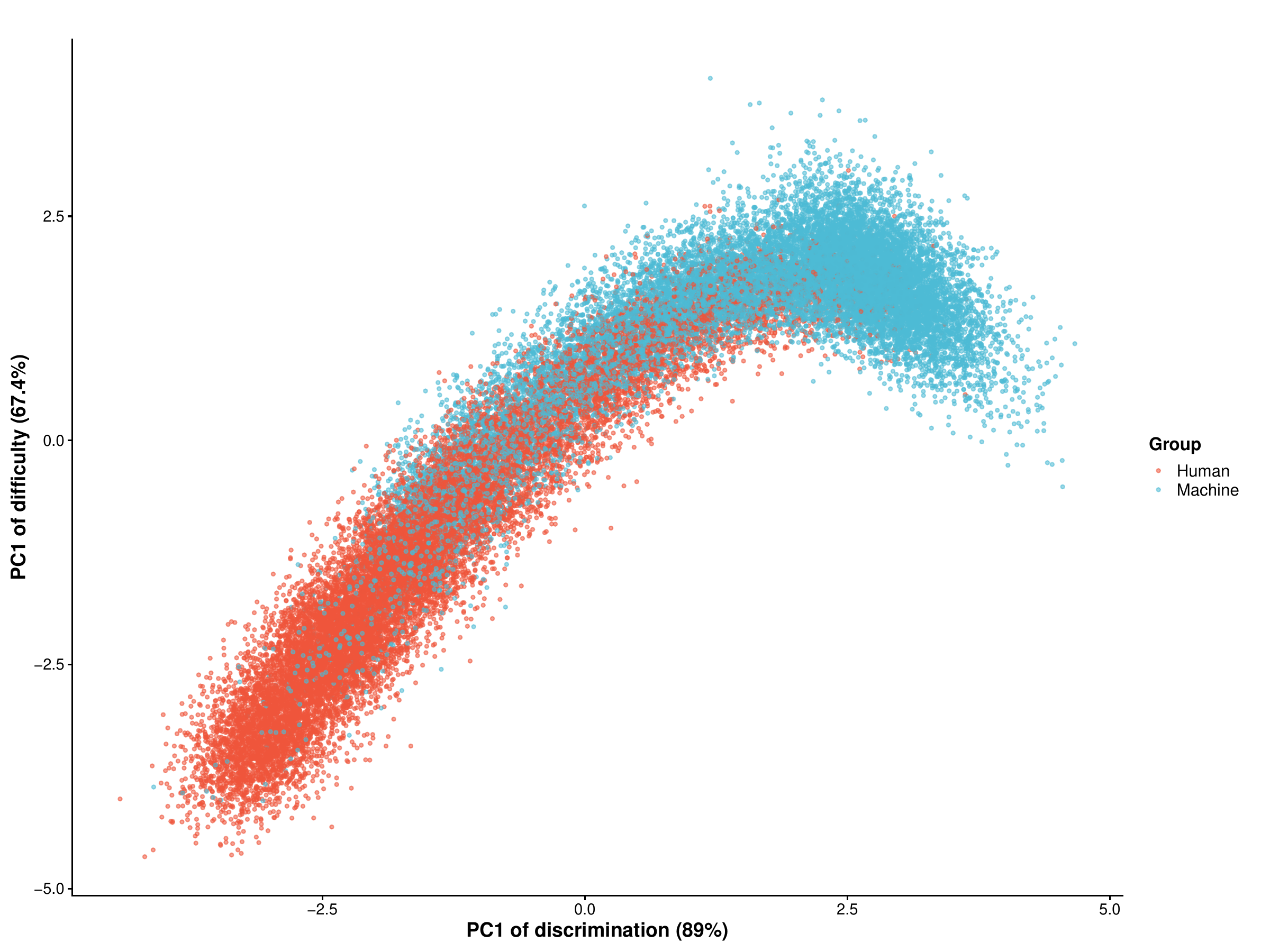}
    \caption{First principal component visualization of the IRT parameter space. Each point represents a text sample, colored by its true group label.}
    \label{fig:alpha_beta_pc}
\end{figure}

Figure \ref{fig:alpha_beta_pc} visualizes the joint IRT parameter space using the first principal components of $\alpha$ and $\beta$. These two components explain 89.0\% and 67.4\% of their variation, respectively, suggesting that the leading principal components provide a compact yet informative low-dimensional representation of the IRT parameter space. Human-written and machine-generated texts show a clear distributional shift but substantial overlap. Machine-generated samples are concentrated toward the upper-right region, whereas human-written samples are more concentrated toward the lower-left region. This pattern suggests that the IRT space captures graded sample-level heterogeneity rather than simply reproducing binary human-machine labels.

The overall $KL(M||H)$ values further quantify these distributional differences and are shown in Table \ref{tab:kl_div}, where $M$ denotes the distribution of machine-generated samples and $H$ denotes the distribution of human-written samples. These values measure the divergence of the machine-generated text distribution from the human-written reference distribution in each latent IRT dimension. In the overall setting, relatively large divergences are observed in several discrimination and difficulty dimensions, including $\alpha_2$, $\alpha_3$, $\beta_1$, and $\beta_5$. This confirms that the human-machine difference is reflected not only in detection difficulty but also in the discriminative value of benchmark samples. The marginal distributions of both $\alpha$ and $\beta$ across all latent dimensions are further shown in Supplementary Figure S1, which provides a dimension-wise view of the distributional differences summarized by the KL divergence values. These results support the view that IRT parameters can serve as quantitative descriptors of benchmark texts, providing sample-level diagnostics of benchmark heterogeneity and complementing conventional detector-level metrics such as accuracy, F1 score, and AUC. 

To formally assess whether the globally estimated IRT parameter space differs across different sample factors, we further conducted a three-factor MANOVA using domain, text length group, and human-machine label as explanatory variables. The response space consists of all estimated item parameters, including both discrimination parameters $\alpha$ and difficulty-location parameters $\beta$. Because the benchmark contains a large number of samples, even small differences may lead to statistically significant p-values. Therefore, we focus more on the relative magnitude of Wilks' $\Lambda$ \citep{wilks1932} when interpreting the MANOVA results. A smaller Wilks’ $\Lambda$ indicates a stronger multivariate effect or greater separation associated with the corresponding factor, whereas a value closer to one indicates a weaker effect. 
The results, presented in Table \ref{tab:manova_global_irt}, show significant main effects for all three factors. Among them, the human-machine group effect is the strongest ($\Lambda=0.372, F=5392.7$), indicating a substantial overall difference between human-written and machine-generated texts in the joint IRT parameter space. The domain effect is also significant ($\Lambda=0.890, F=127.3$), suggesting that benchmark samples from different domains exhibit distinct difficulty–discrimination structures. By comparison, the text length effect is weaker but still significant ($\Lambda=0.970, F=49.1$), indicating that text length modulates the IRT parameter space but is not the dominant source of variation.

For interaction effects, the domain-by-group interaction ($\Lambda=0.832, F=202.6$) suggests that the human-machine difference varies across domains, which is consistent with the domain-specific PCA and KL divergence analyses described in Section \ref{sec:domain}. The length-by-group interaction ($\Lambda=0.960, F=65.1$) further indicates that the human-machine difference is also affected by text length. Although the higher-order interactions are significant, their Wilks' $\Lambda$ values are close to one, suggesting relatively weaker effects compared with the main human-machine group effect and the domain-by-group interaction. Overall, these results confirm that the IRT parameter space captures not only global human-machine differences, but also domain- and length-dependent sample heterogeneity. We therefore further examine the IRT-based parameter space within specific domains and text length groups.

\begin{figure}
    \centering
    \includegraphics[width=0.85\linewidth]{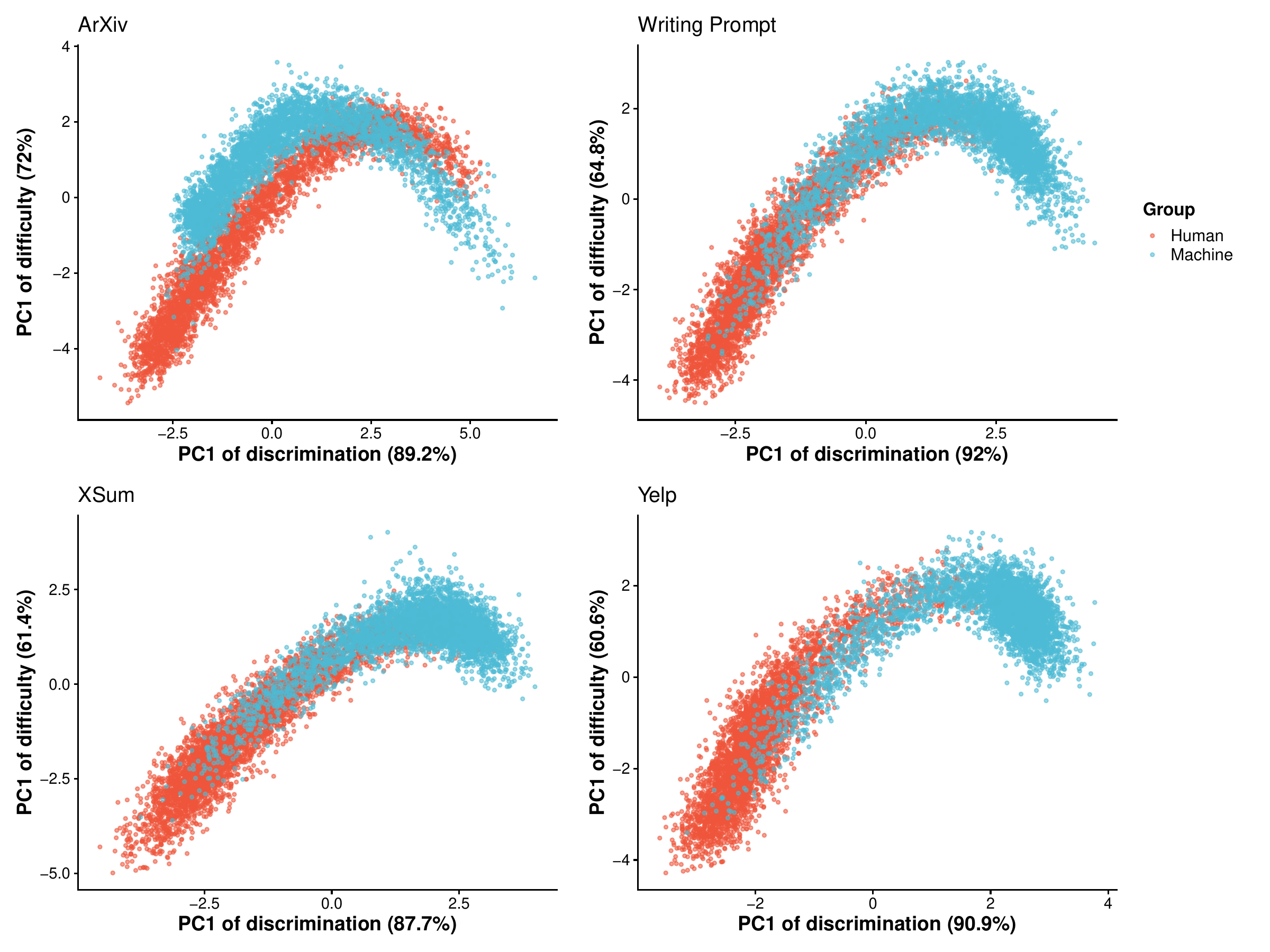}
    \caption{First principal component visualization of the domain-specific IRT parameter space. Each point represents a text sample, colored by its true group label.}
    \label{fig:domain_pc}
\end{figure}

\subsubsection{Domain-specific analysis}\label{sec:domain}
We next examine whether the IRT-based human-machine separation is consistent across domains. Figure \ref{fig:domain_pc} presents the domain-specific IRT representations for ArXiv, Writing Prompt, XSum, and Yelp Reviews using the first principal components of both $\alpha$ and $\beta$. Across all four domains, human-written and machine-generated texts form partially separated but overlapping regions, indicating that the human-machine distinction remains graded within each domain. However, the separation patterns vary across domains. ArXiv shows broadly similar curved trajectories for the two groups, with machine-generated texts concentrated on a narrower portion of the curve. In contrast, Writing Prompt, XSum, and Yelp Reviews exhibit clearer group-level separation, more closely resembling the global multi-domain structure. These visual patterns suggest that the IRT parameter space captures domain-dependent sample heterogeneity rather than a single universal human-machine boundary.

\begin{table*}[!htbp]
\caption{Domain-specific and overall summary of $KL(M||H)$ between machine-generated and human-written text distributions for the estimated IRT discrimination ($\alpha$) and difficulty-location ($\beta$) parameters across five latent dimensions.}
\label{tab:kl_div}

\footnotesize
\setlength{\tabcolsep}{2pt}
\vskip 0.15in
\begin{center}
\begin{sc}
\begin{tabular}[ht]{l|l|rrrrr|rrrrr}
\toprule
&&$\alpha_1$&$\alpha_2$&$\alpha_3$&$\alpha_4$&$\alpha_5$&$\beta_1$&$\beta_2$&$\beta_3$&$\beta_4$&$\beta_5$\\ 
\hline
\multirow{3}{*}{Domain} & ArXiv & 0.289 & 0.221 & 0.073 & 0.899 & 0.126 & 0.668 & 0.640 & 0.180 & 0.149 & 0.700\\
&Writing Prompt & 1.252 & 4.896 & 2.060 & 8.347 & 2.757 & 2.387 & 0.977 & 0.400 & 0.022 & 1.811\\
&XSum & 1.375 & 3.129 & 1.895 & 4.760 & 2.299 & 3.005 & 1.128 & 0.309 & 0.065 & 3.264\\
&Yelp & 2.754 & 13.446 & 2.687 & 15.770 & 6.604 & 3.459 & 1.151 & 0.574 & 0.134 & 1.812\\
\hline
\multirow{3}{*}{Length} &Short&1.395&7.470&2.122&7.918&4.507&2.292&1.010&0.377&0.070&1.561\\
&Medium&0.911&4.585&1.401&6.434&1.418&1.371&0.720&0.319&0.038&1.016\\
&Long&0.994&3.684&1.825&5.598&1.966&1.461&0.721&0.246&0.027&1.126\\
\hline
Overall& &0.821& 7.120 & 2.541& 0.967& 1.104& 2.122& 0.356& 1.019& 0.941& 1.811\\
\bottomrule
\end{tabular}
    \end{sc}
    \end{center}
    \vskip -0.1in
\end{table*}

The domain-specific $KL(M||H)$ values in Table \ref{tab:kl_div} provide quantitative support for the visual patterns. For ArXiv, the divergences are generally moderate with the largest values appearing in $\alpha_4=0.899$, $\beta_5=0.700$, $\beta_1=0.668$, and $\beta_2=0.640$. This indicates that machine-generated academic texts are not strongly separated from human-written academic texts in the marginal IRT parameter distributions. As an auxiliary check, we also compute the reverse divergence $KL(H||M)$, reported in Supplementary Table S1, which suggests a coverage asymmetry: human-written ArXiv texts span a broader region of the difficulty-discrimination space, whereas machine-generated ArXiv texts are more concentrated. From a benchmark evaluation perspective, this asymmetry suggests that balanced class proportions do not necessarily imply comparable coverage of the latent measurement space.

By contrast, Writing Prompt, XSum, and Yelp exhibit much stronger domain-specific divergences, particularly in the discrimination parameters. Writing Prompt shows large KL divergences for $\alpha_4$ (8.347), $\alpha_2$ (4.896), and $\alpha_5$ (2.757), whereas its largest difficulty-related divergences, such as $\beta_1$ (2.387) and $\beta_5$ (1.811), are smaller. XSum shows large divergences in $\alpha_4$ (4.760), $\alpha_2$ (3.129), and $\alpha_5$ (2.299), together with substantial difficulty-related divergences in $\beta_5$ (3.264) and $\beta_1$ (3.005). Yelp Reviews exhibits the strongest separation among all domains, with particularly large divergences in $\alpha_4$ (15.770), $\alpha_2$ (13.446), $\alpha_5$ (6.604), $\beta_1$ (3.459), and $\beta_5$ (1.812). These results indicate that domain-specific human-machine differences are reflected primarily in the discrimination-related dimensions, although difficulty-related dimensions also contribute in some domains. Overall, the domain-level analysis shows that IRT parameters capture both shared and domain-specific measurement properties of benchmark texts under detector evaluation. The dimension-wise distributions of the estimated $\alpha$ and $\beta$ parameters for each domain are shown in Supplementary Figures S2-S5, providing a more detailed view of the domain-specific patterns summarized by the KL divergence values.

\subsubsection{Length-stratified analysis}\label{sec:length}

\begin{figure}
    \centering
    \includegraphics[width=1\linewidth]{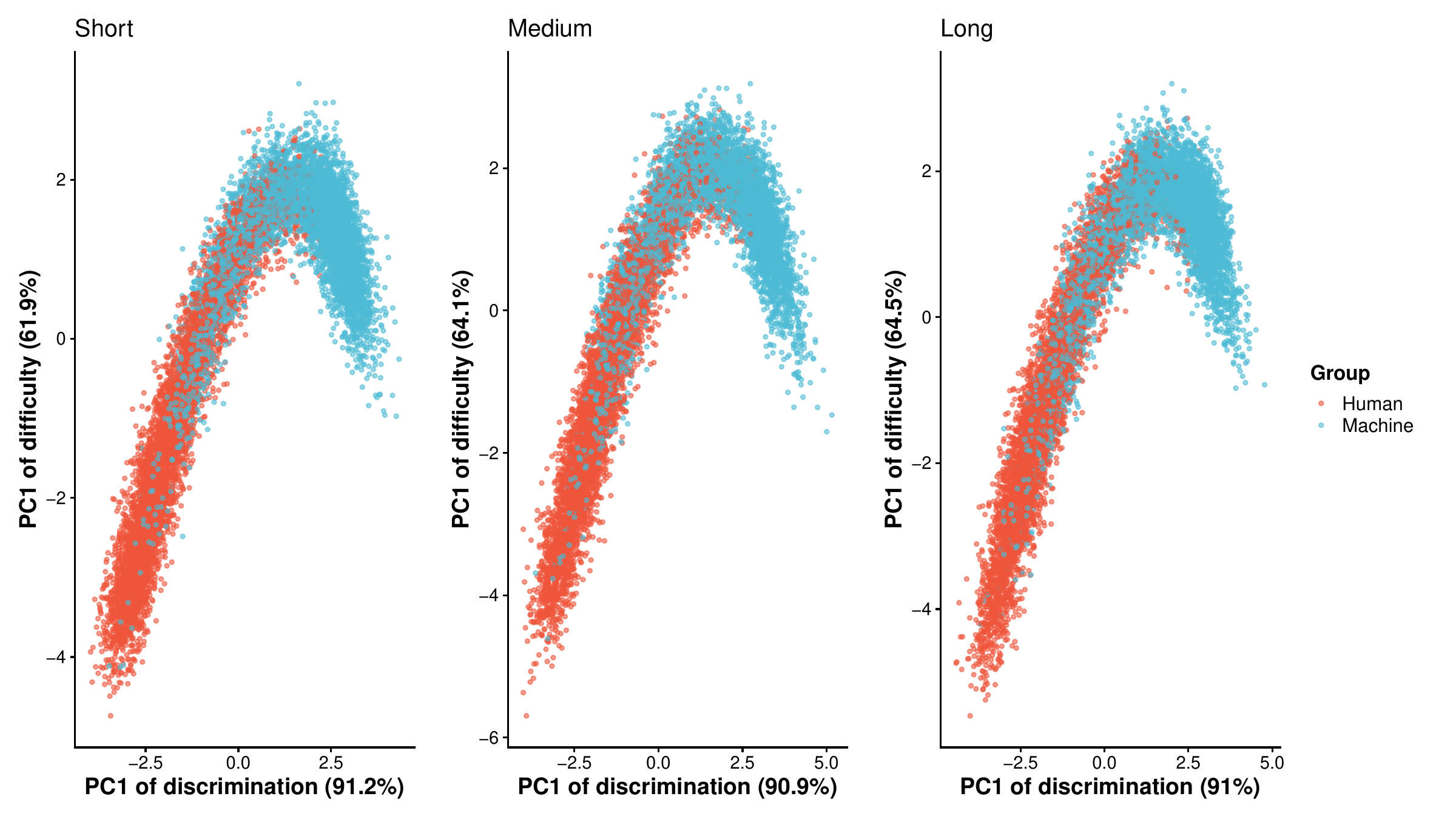}
    \caption{First principal component visualization of the text length stratified IRT parameter space. Each point represents a text sample, colored by its true group label.}
    \label{fig:pc_length}
\end{figure}

We further examine whether the IRT-based human-machine separation varies across text lengths. The benchmark samples are divided into short, medium, and long groups, as summarized in Table \ref{tab:length}. The first principal component visualization of the discrimination and difficulty-location parameters is shown in Figure \ref{fig:pc_length}. The three panels show broadly similar curved structures in the IRT parameter space. Across all length groups, human-written and machine-generated texts are partially separated but still overlap. Machine-generated samples are more concentrated toward the upper-right region of the IRT space, whereas human-written samples extend more strongly toward the lower-left branch. These results indicate that the human-machine differences captured by the IRT parameters are not simply associated with text length, as similar separation patterns persist within each length stratum.

The length-specific $KL(M||H)$ values in Table \ref{tab:kl_div} further show that the main distributional differences are concentrated in the discrimination dimensions. For short texts, the largest divergences occur in $\alpha_4$ (7.918), $\alpha_2$ (7.470), and $\alpha_5$ (4.507). Medium texts show a similar pattern, with large divergences in $\alpha_4$ (6.434) and $\alpha_2$ (4.585), while long texts also exhibit the largest divergences in $\alpha_4$ (5.598), $\alpha_2$ (3.684), and $\alpha_5$ (1.966). By contrast, the difficulty-related divergences are generally smaller across all three groups. These results suggest that the strength of human-machine separation varies with text length, with short texts showing the strongest discrimination-based divergence. Overall, the length-stratified analysis confirms that IRT parameters capture sample-level heterogeneity beyond simple length effects. From a benchmark design perspective, these findings suggest that text length should be considered alongside class and domain composition when assessing the measurement properties of an evaluation benchmark.

\begin{table*}

    \centering\caption{Distribution of benchmark samples across text length groups.}\label{tab:length}
    \begin{tabular}{@{}lcccc@{}}
    \toprule
       Category& Length (words)  & Machine & Human& Total\\
       \midrule
       Short &$\leq 150 $ & 5588  & 5390&10978 \\
       Medium &$[151,250]$ & 5301 & 5392&10693\\
       Long &$\geq 251$& 5111& 5218&10329\\
       \bottomrule       
    \end{tabular}
\end{table*}

\subsection{Multidimensional comparison of detector abilities}
After analyzing the item-level parameters $\alpha$ and $\beta$, we further examine the detector-level ability parameters $\theta$. The purpose of this analysis is not to replace conventional metrics such as AUC, F1 score, or accuracy, but to provide a latent representation of detector response patterns in the IRT space for comparing heterogeneous detection approaches. While $\alpha$ and $\beta$ characterize the difficulty and discriminative value of benchmark texts, $\theta$ provides a complementary view of how different detectors behave in the latent IRT space. Therefore, $\theta$ links detector-level performance to the underlying detector-text response structure and helps explain why detectors may behave differently across domains and evaluation metrics. Table \ref{tab:theta} reports the detector ability estimates from the five-dimensional IRT model in the multi-domain setting after varimax orthogonal rotation. We apply the rotation to obtain a more interpretable orientation of the latent ability space. 

\begin{figure}
    \centering
    \includegraphics[width=0.85\linewidth]{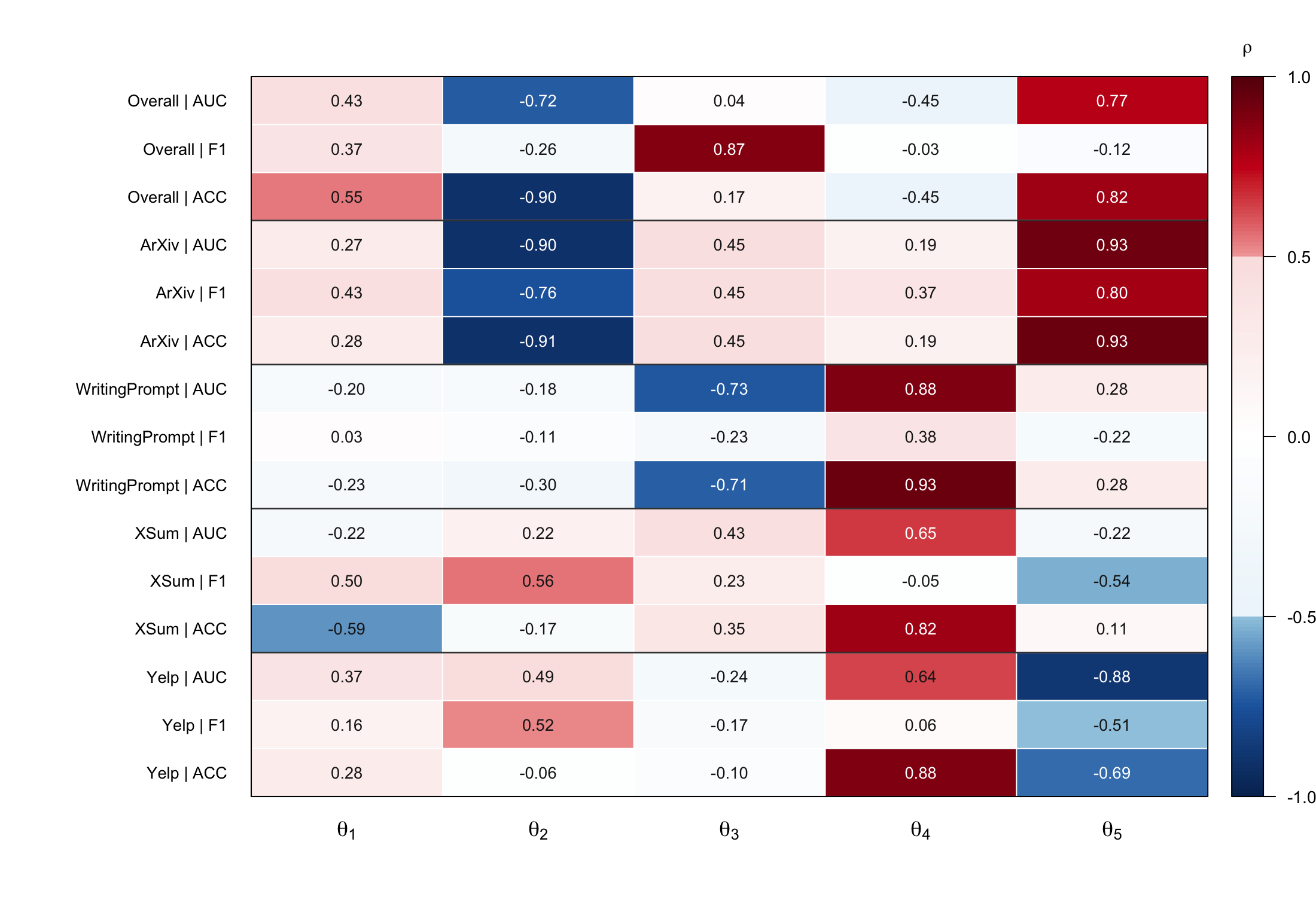}
    \caption{Spearman correlation heatmap between detector ability estimates ($\theta$) and detection performance metrics across overall and domain-specific benchmarks. Rows represent benchmark-metric pairs, columns represent latent IRT dimensions, and colors indicate the Spearman correlation coefficient $\rho$. The heatmap shows that different domains and metrics align with different ability dimensions. For domain-specific rows, the dimensions refer to the corresponding within-domain latent spaces and are not directly aligned across domains.}
    \label{fig:rho_performance}
\end{figure}

The results show that different latent dimensions capture different aspects of detector behavior. The third dimension is the dominant common dimension: all detectors have their largest ability estimates on $D_3$, indicating that this dimension captures the major source of shared variation across detectors. This interpretation is further supported by the correlation analysis in Figure \ref{fig:rho_performance}, where $\theta_3$ shows a strong positive Spearman correlation with the overall F1 score ($\rho=0.87$). Thus, $\theta_3$ can be interpreted as a general ability factor that is particularly aligned with threshold-dependent detection performance.
In contrast, $\theta_5$ shows a distinct pattern across detector types. Zero-shot or score-based detectors, such as DetectGPT, Entropy, Likelihood, LogRank, and DNA-GPT, have positive ability estimates on $\theta_5$, whereas supervised detectors such as RoBERTa-Base, RoBERTa-Large, and RoBERTa-ChatGPT have negative estimates. This pattern suggests that $\theta_5$ captures a secondary axis that separates score-based methods from supervised classification-based methods. Moreover, $\theta_5$ is positively associated with overall AUC ($\rho=0.77$) and accuracy ($\rho=0.82$), suggesting that this dimension is related to the detectors' overall ranking and classification performance. By comparison, $\theta_2$ is negatively correlated with overall AUC ($\rho=-0.72$) and accuracy ($\rho=-0.90$), indicating that detectors with smaller values on this rotated dimension tend to achieve stronger overall ranking and classification performance. Although $\theta_2$ represents a weaker dimension than $\theta_3$, it still captures non-negligible variation, with relatively larger values for RoBERTa-based detectors. The remaining dimensions, $\theta_1$ and $\theta_4$, are generally close to zero for most detectors in the multi-domain setting.
Overall, the rotated ability matrix indicates that detector performance is better understood as a multidimensional structure, consisting of one dominant common ability dimension and several weaker dimensions that distinguish detector types and evaluation behaviors.

\begin{table*}[t]
\centering
\caption{Rotated 
$\theta$ estimates for all detectors from the five-dimensional IRT model. 
A dash indicates that the corresponding absolute value is below 0.015.} \label{tab:theta}%
\vskip 0.15in
    \begin{tabular}{@{}lccccc@{}}
    \toprule
    \multicolumn{1}{c}{\textbf{Detector}}      & $\theta_1$ & $\theta_2$& $\theta_3$ & $\theta_4$& $\theta_5$ \\
          \midrule
    Binoculars  & -  & 0.02  & 1.89  & -  & 0.02 \\
    DetectGPT & - & -0.03 & 0.49  & -  & 0.32 \\
    DNA-GPT & 0.02  & 0.03  & 0.91  & -  & 0.16 \\
    Entropy & -  & 0.03  & 0.84  & - & 0.25 \\
    Fast-DetectGPT & - & 0.02  & 2.00  & - & 0.03 \\
    Likelihood & 0.02  & 0.02  & 1.70  & -  & 0.15 \\
    LogRank & 0.02  & - & 1.58  & - & 0.22 \\
    LRR   & -  & 0.04  & 1.87  & -  & - \\
    NPR   & - & 0.06  & 1.19  & - & 0.03 \\
    Rank  & - & 0.04  & 1.76  & -  & 0.02 \\
    Lastde & 0.02  & 0.03  & 1.89  & - & 0.04 \\
    RoBERTa-Base & -  & 0.09  & 1.30  & - & -0.12 \\
    RoBERTa-Large & - & 0.10  & 1.23  & 0.02  & -0.12 \\
    RoBERTa-ChatGPT & -  & 0.12  & 0.92  & - & -0.08 \\
    \bottomrule
    \end{tabular}%

\end{table*}%

\begin{table}[t]
  \caption{Domain-specific detector ability estimates ($\theta$) under the five-dimensional IRT model. The ability parameters are re-estimated separately within each domain. Rows correspond to detectors, and columns report the estimated ability values for each latent dimension within ArXiv, Writing Prompt, XSum, and Yelp Reviews. A dash indicates that the corresponding absolute value is below 0.015.}  \label{tab:theta_domain}
    \begin{tabular}{@{}l|ccccc|ccccc@{}}
      \toprule
          & \multicolumn{5}{c}{ArXiv}      & \multicolumn{5}{c}{Writing Prompt} \\
          \midrule
          & $\theta_1$ & $\theta_2$ & $\theta_3$ & $\theta_4$ & $\theta_5$ & $\theta_1$ & $\theta_2$ & $\theta_3$ & $\theta_4$ & $\theta_5$ \\
          \midrule
    Binoculars    & -0.09 & -1.82 & -  & -0.05 & -0.21 & -0.03 & -2.30 & -0.22 & - & - \\
    DetectGPT     & 0.13  & -1.26 & -  & -0.02 & -0.21 & -  & -1.20 & -0.42 & -  & - \\
    DNA-GPT    & - & -1.87 & -  & - & -0.19 & 0.02  & -2.38 & -0.15 & -0.06 & - \\
    Entropy    & 0.10  & -2.18 & -  & -0.02 & -0.12 & - & -1.91 & -0.07 & -0.14 & - \\
    Fast-DetectGPT     & -0.08 & -1.90 & -  & -0.05 & -0.23 & - & -2.69 & -0.20 & -  & - \\
    Likelihood     & -  & -3.14 &-  & -  & -0.03 & -  & -2.64 & -0.09 & -0.02 & -0.02 \\
    LogRank    & -0.03 & -3.30 & - & - & -0.02 & -  & -0.49 & -0.49 & -0.03 & - \\
    LRR     & -  & -2.38 & -  & - & -0.12 & -  & -2.63 & -  & -0.10 & - \\
   NPR     & 0.12  & -2.20 & -  & -  & -0.12 & - & -0.82 & -0.15 & -0.24 & - \\
     Rank    & 0.08  & -2.61 & 0.04  & -  & -0.07 & -  & -2.48 & 0.02  & -0.12 & - \\
    Lastde   & -0.03 & -1.93 & -  & -0.04 & -0.18 & -0.03 & -2.51 & -0.19 & -  &- \\
    RoBERTa-Base    & -  & -0.02 & -  & - & -0.38 & 0.02  & -2.41 & -  & -0.15 & - \\
    RoBERTa-Large   & 0.02  & -0.08 & - & -  & -0.36 & -  & -2.06 & -  & -0.18 & -0.03 \\
    RoBERTa-ChatGPT    & 0.03  & -0.17 & -  & -  & -0.35 & -  & -1.77 & -0.08 & -0.17 & - \\
    \bottomrule
          & \multicolumn{5}{c}{XSum}    & \multicolumn{5}{c}{Yelp}   \\
          \midrule
          & $\theta_1$ & $\theta_2$ & $\theta_3$ & $\theta_4$ & $\theta_5$ & $\theta_1$ & $\theta_2$ & $\theta_3$ & $\theta_4$ & $\theta_5$ \\
          \midrule
       Binoculars   & -  & -2.06 & -0.02 & 0.02  & -0.06 & -  & -3.28 & -  & 0.07  & -0.12 \\
        DetectGPT  & -  & -1.00 & - & 0.06  & -0.24 & -  & -0.02 & -  & -0.02 & -0.41 \\
        DNA-GPT  & -  & 0.14  & -  & -  & -0.43 & - & -1.79 & -0.02 & 0.02  & -0.27 \\
        Entropy  & - & -1.03 & - & 0.12  & -0.23 & -  & -0.30 & -  & 0.02  & -0.41 \\
        Fast-DetectGPT  & -  & -2.13 & -  & -  & -0.07 & - & -3.50 & - & 0.07  & -0.11 \\
        Likelihood  & -  & -1.27 & -  & 0.10  & -0.18 & 0.02  & -2.62 & - & 0.09  & -0.18 \\
        LogRank  & -  & -2.03 & -  & 0.07  & -0.06 & 0.02  & -3.68 & -  & 0.05  & -0.04 \\
        LRR  &-  & -1.99 & -  & 0.03  & -0.06 & -  & -3.60 & - & -0.02 & - \\
       NPR   & 0.02  & -1.33 & -0.03 & -  & -0.17 & -  & -3.06 & -  & -0.15 & - \\
       Rank   & - & -1.68 & -  & 0.03  & -0.11 & -0.03 & -3.24 & - & -0.08 & - \\
       Lastde   & - & -1.97 & - & -  & -0.07 & -  & -3.26 & -  & 0.08  & -0.13 \\
       RoBERTa-Base   & 0.03  & -1.82 & -  & -0.04 & -0.09 & - & -3.05 & -  & -0.09 & -0.06 \\
       RoBERTa-Large   & 0.05  & -1.74 & -  & -0.02 & -0.09 & -  & -3.21 & - & -0.12 & - \\
       RoBERTa-ChatGPT   & -  & -0.62 & -0.02 & -0.05 & -0.28 & - & -3.05 & - & -0.03 & -0.09 \\
          \bottomrule
    \end{tabular}%

\end{table}%

Table \ref{tab:theta_domain} further reports the domain-specific detector ability estimates. These parameters are re-estimated separately within each domain. Therefore, we focus on relative patterns within each domain rather than making direct numerical comparisons across domains. Because the domain-specific models are fitted separately, the latent dimension labels are specific to each fitted model and should not be interpreted as directly aligned across domains. The correlation patterns in Figure \ref{fig:rho_performance} show that different domains emphasize different latent ability dimensions. In ArXiv, the strongest variation is concentrated on $\theta_2$, where many zero-shot and score-based detectors show large negative values, while RoBERTa-based detectors are closer to zero. This suggests that academic writing induces a distinct ability structure, with detector differences mainly organized along a domain-specific dimension. Consistently, $\theta_2$ is strongly negatively correlated with AUC ($\rho=-0.90$), F1 score ($\rho=-0.71$), and accuracy ($\rho=-0.91$), whereas $\theta_5$ is strongly positively correlated with AUC ($\rho=0.93$), F1 score ($\rho=0.80$), and accuracy ($\rho=0.93$). These results indicate that ArXiv exhibits a relatively stable ability-performance structure, in which the latent dimensions align well with conventional performance metrics.

In contrast, Writing Prompt, XSum, and Yelp Reviews exhibit more heterogeneous ability-performance relationships. For Writing Prompt, the ability variation is more distributed across $\theta_2$, $\theta_3$, and $\theta_4$. Among them, $\theta_4$ is strongly correlated with AUC and accuracy ($\rho=0.88$ and $0.93$), whereas $\theta_3$ is negatively correlated with these metrics. This suggests that the ability dimension most relevant to creative writing differs from the dominant common dimension observed in the multi-domain setting. For XSum and Yelp Reviews, the ability patterns are mainly associated with $\theta_2$, $\theta_4$, and $\theta_5$, differing from both ArXiv and Writing Prompt. In XSum, $\theta_4$ is positively correlated with AUC and accuracy ($\rho=0.65$ and $0.82$), while $\theta_2$ and $\theta_5$ also contribute to F1 score. Yelp Reviews shows another distinct pattern. AUC and accuracy are positively associated with $\theta_4$ ($\rho=0.64$ and $0.88$), but negatively associated with $\theta_5$ ($\rho=-0.88$ and $-0.69$). These results indicate that detector ability is not fixed across domains. Instead, each domain is associated with a different latent ability structure, and different evaluation metrics may align with different IRT dimensions. Together, Tables \ref{tab:theta} and \ref{tab:theta_domain} and Figure \ref{fig:rho_performance} show that the IRT framework provides a multidimensional representation of detector ability, enabling comparisons of heterogeneous detection approaches beyond aggregate performance metrics and revealing domain-dependent detector response patterns.

\subsection{Compact benchmark construction}
As a practical use of the estimated IRT item parameters, we further examine whether the IRT representation can support compact benchmark construction. Using $E_i=(\hat{\alpha}_i,\hat{\beta}_i)$, we apply $K$-means clustering and select texts closest to the cluster centers as representative samples. The objective is to construct a more compact evaluation benchmark that preserves the main comparative detector-performance patterns of the full benchmark, rather than to improve detector performance.

\begin{figure}
    \centering
    \includegraphics[width=0.7\linewidth]{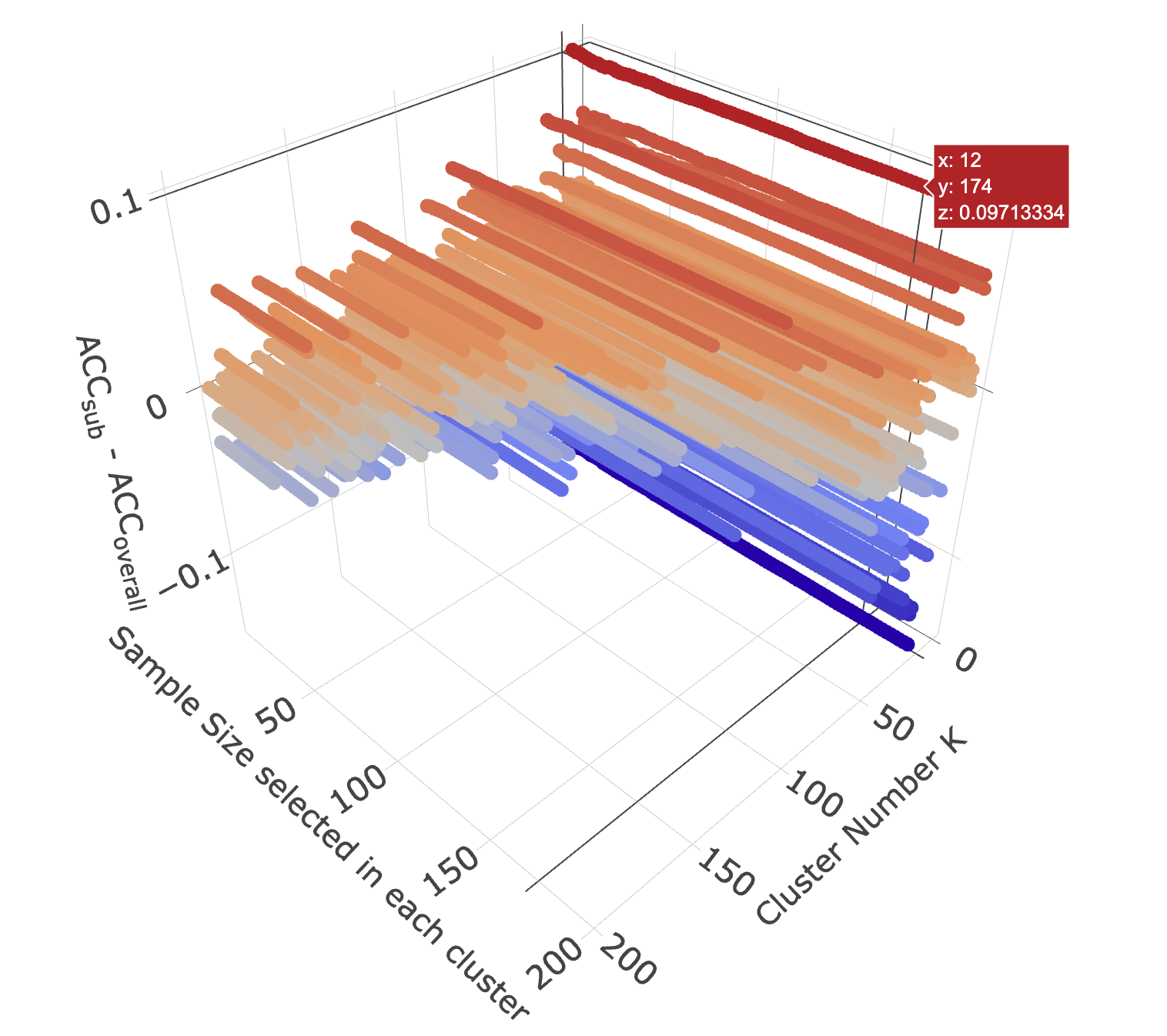}
    \caption{Difference between subset-based accuracy and full-benchmark accuracy under different values of $K$ and $n_i$. Colors indicate the magnitude of $\Delta \mathrm{ACC}$.}
    \label{fig:acc_improve}
\end{figure}

With $K=12$ and $n_i=174$, the resulting subset contains 2,088 texts and retains samples from all four domains and both human-written and machine-generated classes (Table \ref{tab:selected}). As shown in Figure \ref{fig:acc_improve}, the subset-based accuracies follow a similar detector-wise performance pattern to the full benchmark, although the absolute accuracies are higher for several detectors. These results suggest that the IRT-guided subset can provide a compact approximation to the full evaluation set in terms of comparative accuracy patterns, although differences in absolute accuracy remain. Additional sensitivity analyses over different choices of $K$ and $n_i$ are reported in Supplementary Figure S6.

\begin{table*}[t]
  \centering
  \caption{Distribution of the selected subsample across domains and classes under the clustering-based subsampling scheme with $K=12$ and $n_i=174$.}\label{tab:selected}
    \begin{tabular}{@{}lcc@{}}
    \toprule
      Domains    & \multicolumn{1}{c}{Human} & \multicolumn{1}{c}{Machine} \\
          \midrule
    ArXiv  & 258 &191 \\
    Writing Prompt     & 324&190 \\
    XSum    & 324 &250\\
    Yelp  & 337 &214\\
    \midrule
    All  & 1243&845 \\
    \bottomrule
    \end{tabular}%
\end{table*}%

\section{Discussion}
\subsection{Implications for benchmarking LLM-generated text detectors}
The results highlight the importance of considering benchmark composition when comparing LLM-generated text detection technologies. As shown by the conventional performance metrics, detector rankings vary across domains and evaluation criteria. The multidimensional IRT representation provides a complementary perspective by characterizing detector response patterns beyond aggregate performance rankings. Together, these findings suggest that systematic detector benchmarking should consider not only overall performance but also how different detectors respond to heterogeneous evaluation texts.

\subsection{Implications for benchmark design}
The estimated IRT item parameters also provide insights into the measurement properties of evaluation benchmarks. The observed differences across human- and machine-generated texts, domains, and text lengths suggest that balanced sample counts and domain diversity do not necessarily imply comparable coverage of the latent measurement space. From this perspective, benchmark evaluation can be complemented by considering text-level difficulty and discrimination, as well as the coverage of heterogeneous detector-response patterns. The compact benchmark experiment further illustrates the potential of IRT-based representations for constructing smaller evaluation sets while retaining the main comparative accuracy patterns of the full benchmark.

\subsection{Generalizability and limitations}
More broadly, the results motivate a shift from detector-centered performance comparison toward measurement-based evaluation of the entire detector–benchmark system. The proposed framework is not tied to a particular detector architecture or benchmark domain and can, in principle, be applied to other detector collections and future evaluation settings. Several limitations remain, including the reliance on thresholded detector outputs, the relative interpretation of difficulty-location parameters within the fitted latent space, and the omission of factors such as prompting, decoding, post-editing, and generator characteristics. Moreover, the present analysis is based on 14 detectors, and the ability of the compact benchmark to preserve evaluation patterns for previously unseen detectors remains to be investigated.

\section{Conclusion}

In this paper, In this paper, we propose a measurement-based benchmarking framework for LLM-generated text detection using IRT. By treating benchmark texts as items and detectors as respondents, the framework jointly characterizes text-level measurement properties and multidimensional detector response patterns, complementing conventional aggregate performance metrics such as accuracy, F1 score, and AUC. 

Experiments on 32K human-written and machine-generated texts evaluated by 14 detectors support this view. The two groups occupy distinct but overlapping regions of the IRT item-parameter space, indicating a graded rather than completely separable structure. The human–machine label shows the strongest multivariate association with this space, while domain and text length further modulate benchmark heterogeneity. Detector abilities exhibit multidimensional and domain-dependent patterns, revealing response characteristics that are not fully reflected in aggregate performance metrics. As a practical consequence, the IRT representation also supports compact benchmark construction, with a subset of 2,088 texts largely preserving the comparative accuracy patterns observed on the full benchmark.

\backmatter

\section*{Data Availability} The selected benchmark data and code are available at https://github.com/YbinZhao/text-detector-evaluation.

\section*{Acknowledgements}
This work was partially supported by Shanghai Polytechnic University under Grant No. EGD26QD18 and the National Natural Science Foundation of China (NSFC) under Grant No. 12201158.

\bibliography{ref}

\newpage
\section*{Author Contributions}
Yanbin Zhao conceived the study, developed the main research ideas, supervised the overall direction of the work, and led the manuscript preparation. Zhipeng Chen was responsible for data collection and detector experiments. Guangsheng Bao contributed to the discussion, interpretation of the results, and revision of the manuscript. Guandong Zhao contributed to manuscript revision and editing. Yan Li was responsible for the statistical analysis and contributed to the methodology and manuscript writing. All authors participated in the study, discussed the results, reviewed the manuscript, and approved the final version.
\end{document}


\title{Supplementary Information of ``A Measurement Perspective on Benchmarking
LLM-Generated Text Detectors using Item
Response Theory"}

\author{\fnm{Yanbin} \sur{Zhao}}

\author{\fnm{Zhipeng} \sur{Chen}}

\author{\fnm{Guangsheng} \sur{Bao}}

\author{\fnm{Guandong} \sur{Zhang}}

\author{\fnm{Yan} \sur{Li}}


\maketitle

\setcounter{figure}{0}
\setcounter{table}{0}

\renewcommand{\thefigure}{S\arabic{figure}}
\renewcommand{\thesection}{S\arabic{section}}
\renewcommand{\thetable}{S\arabic{table}}

\section{Additional Distributions of IRT Item Parameters}\label{secA1}

\begin{figure}[!htbp]
    \centering
    \includegraphics[width=1\linewidth]{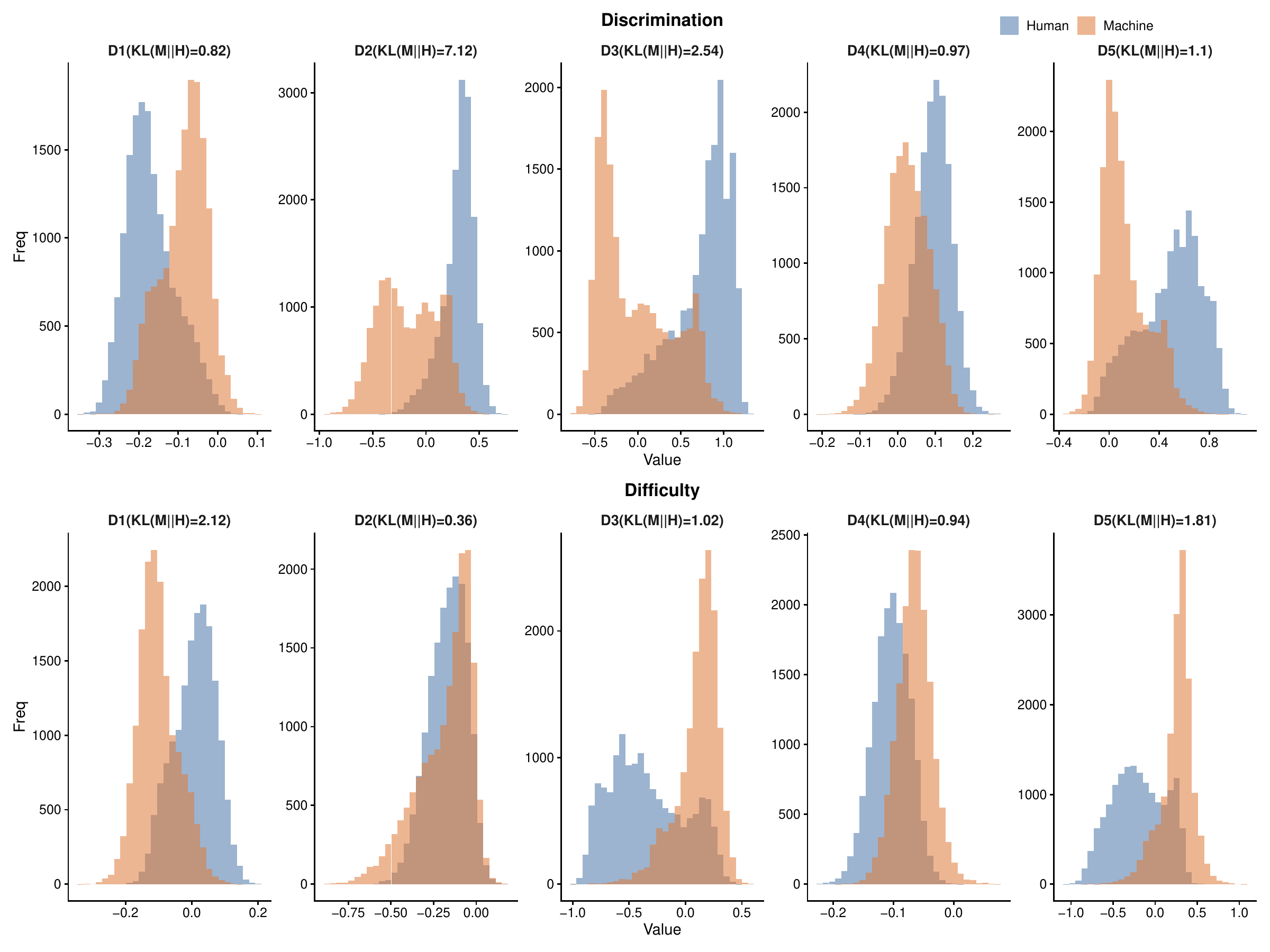}
    \caption{Marginal distributions of the estimated discrimination parameters $\alpha$ and difficulty-related location parameters $\beta$ for human-written and machine-generated texts across the five latent dimensions. Each panel reports the corresponding KL divergence $KL(M||H)$, which measures the divergence of the machine-generated distribution from the human-written reference distribution.}
    \label{fig:alpha_kl}
\end{figure}

In the main text, we provide a principal component visualization and KL divergence analysis to summarize the global structure of the estimated IRT item parameters. To provide a more detailed dimension-wise view, Figure \ref{fig:alpha_kl} presents the marginal distributions of the discrimination parameters $\alpha$ and the difficulty-related location parameters $\beta$ for human-written and machine-generated texts across all five latent dimensions.

The KL divergence is defined as
$$
KL(M\|H)=\sum_x p_M(x)\log\frac{p_M(x)}{p_H(x)},
$$
where $p_M(x)$ and $p_H(x)$ denote the empirical marginal distributions of the machine-generated and human-written groups, respectively. Larger values indicate greater distributional divergence between the two groups along the corresponding IRT dimension, whereas smaller values indicate greater distributional overlap. Since KL divergence is asymmetric, $KL(M\|H)$ specifically measures how the machine-generated distribution diverges from the human-written distribution, which is treated as the reference. We therefore use $KL(M\|H)$ to quantify the divergence of the machine-generated distribution from the human-written reference distribution.

The dimension-wise distributions show that the differences between human-written and machine-generated texts are not uniformly distributed across the IRT parameter space. For the discrimination parameters, the largest divergences are observed in $\alpha_2$ and $\alpha_3$, indicating that these dimensions contribute most strongly to the marginal distributional differences between the two groups. For the difficulty-related location parameters, $\beta_1$ and $\beta_5$ exhibit relatively large divergences, whereas $\beta_2$ shows substantially greater distributional overlap between the two groups. These results provide additional support for the global IRT analysis presented in the main text, showing that the distinction between human-written and machine-generated texts is concentrated in specific discrimination and difficulty-related dimensions rather than being uniformly distributed across the latent space.


\FloatBarrier
\section{Additional Domain-Specific IRT Results}\label{secA2}%
\subsection{Dimension-wise domain-specific item parameter distributions}
\begin{figure}[h]
     \centering
    \includegraphics[width=1\linewidth]{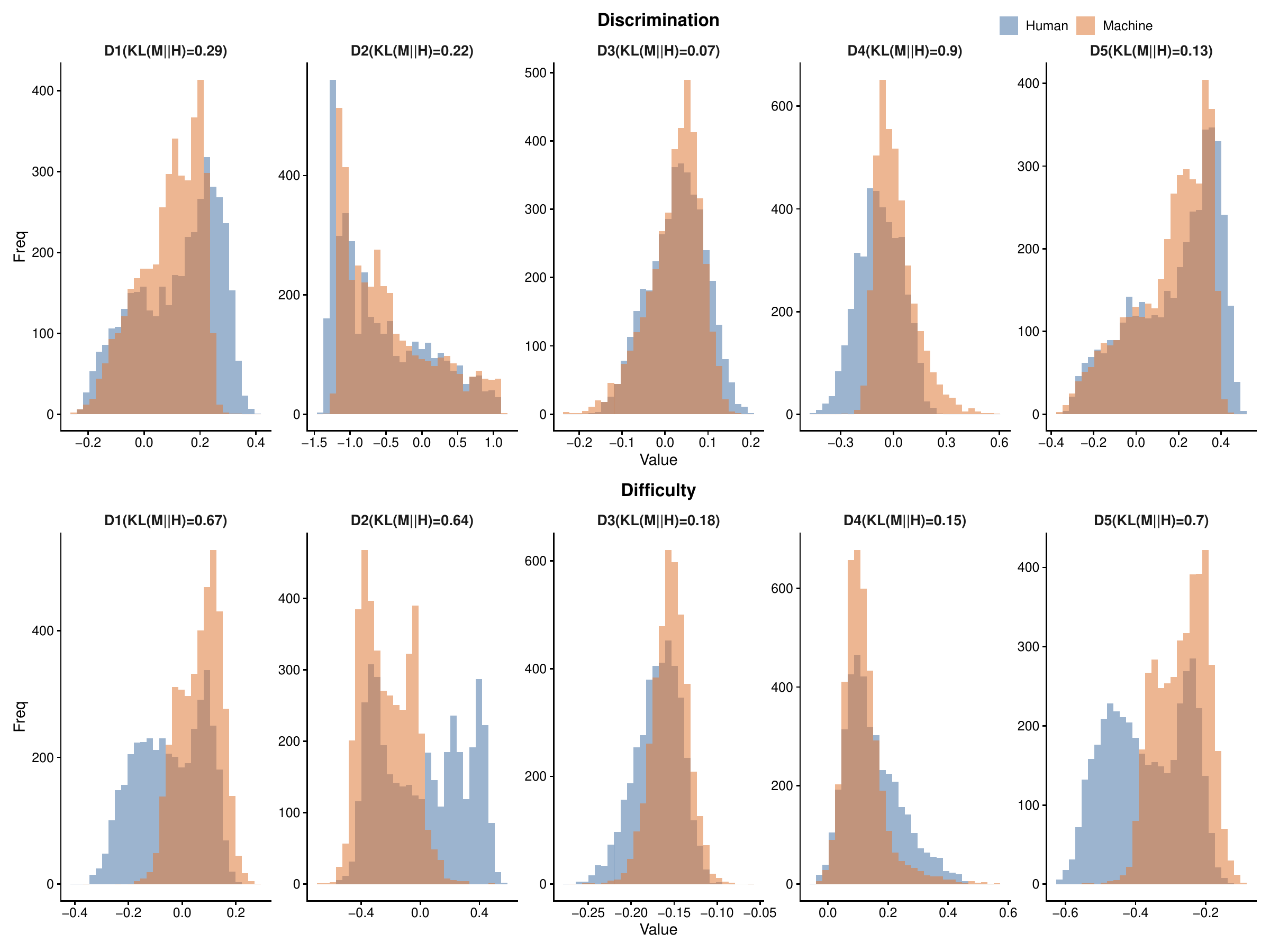}
    \caption{Marginal distributions of the estimated discrimination parameters $\alpha$ and difficulty-related location parameters $\beta$ for human-written and machine-generated texts in the ArXiv domain across the five latent dimensions. Each panel reports the corresponding domain-specific KL divergence $KL(M||H)$.}
    \label{fig:arxiv}
\end{figure}

\begin{figure}[h]
     \centering
    \includegraphics[width=1\linewidth]{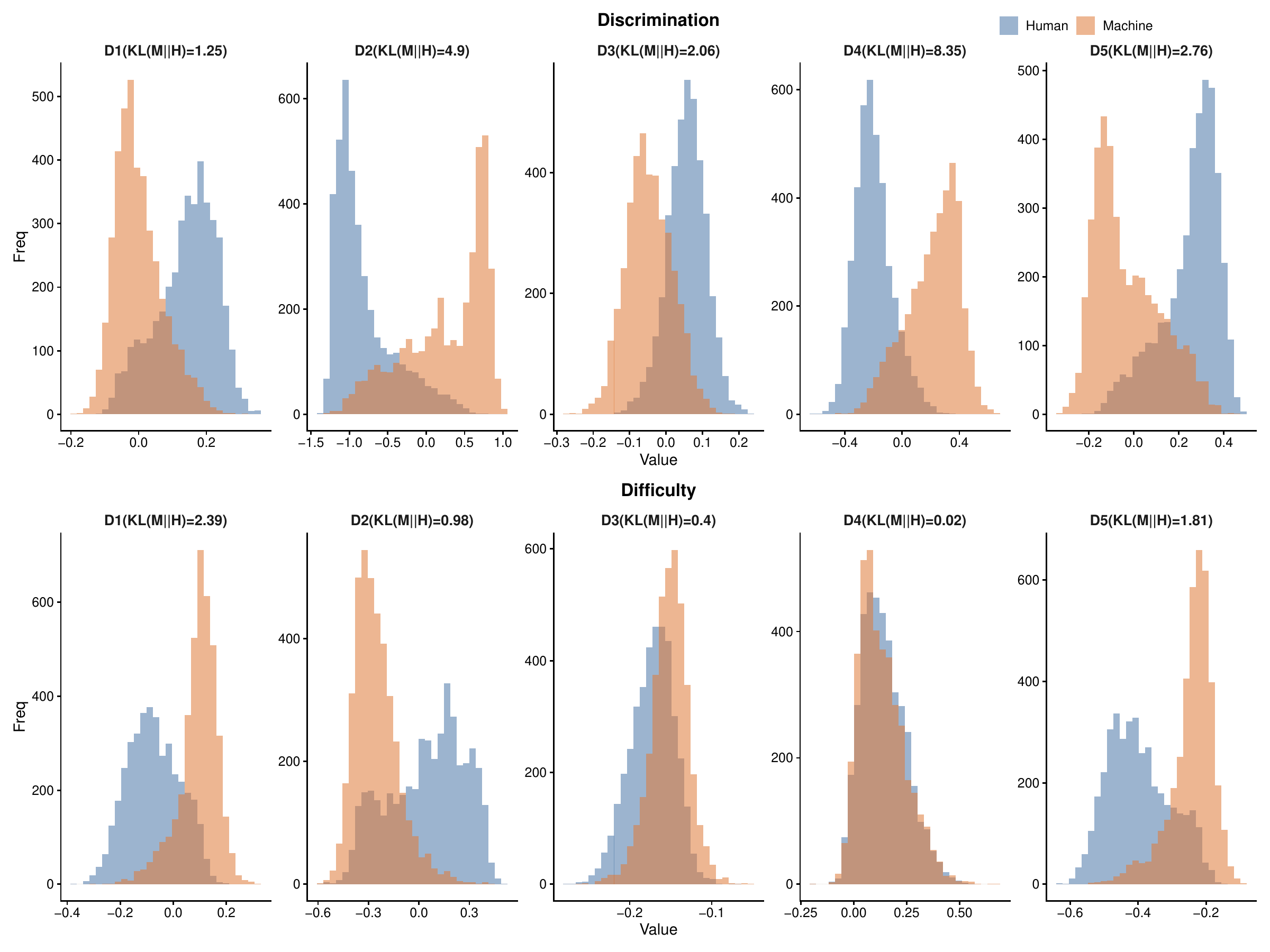}
    \caption{Marginal distributions of the estimated discrimination parameters $\alpha$ and difficulty-related location parameters $\beta$ for human-written and machine-generated texts in the Writing Prompt domain across the five latent dimensions. Each panel reports the corresponding domain-specific KL divergence $KL(M||H)$.}
    \label{fig:writing}
\end{figure}

\begin{figure}[h]
     \centering
    \includegraphics[width=1\linewidth]{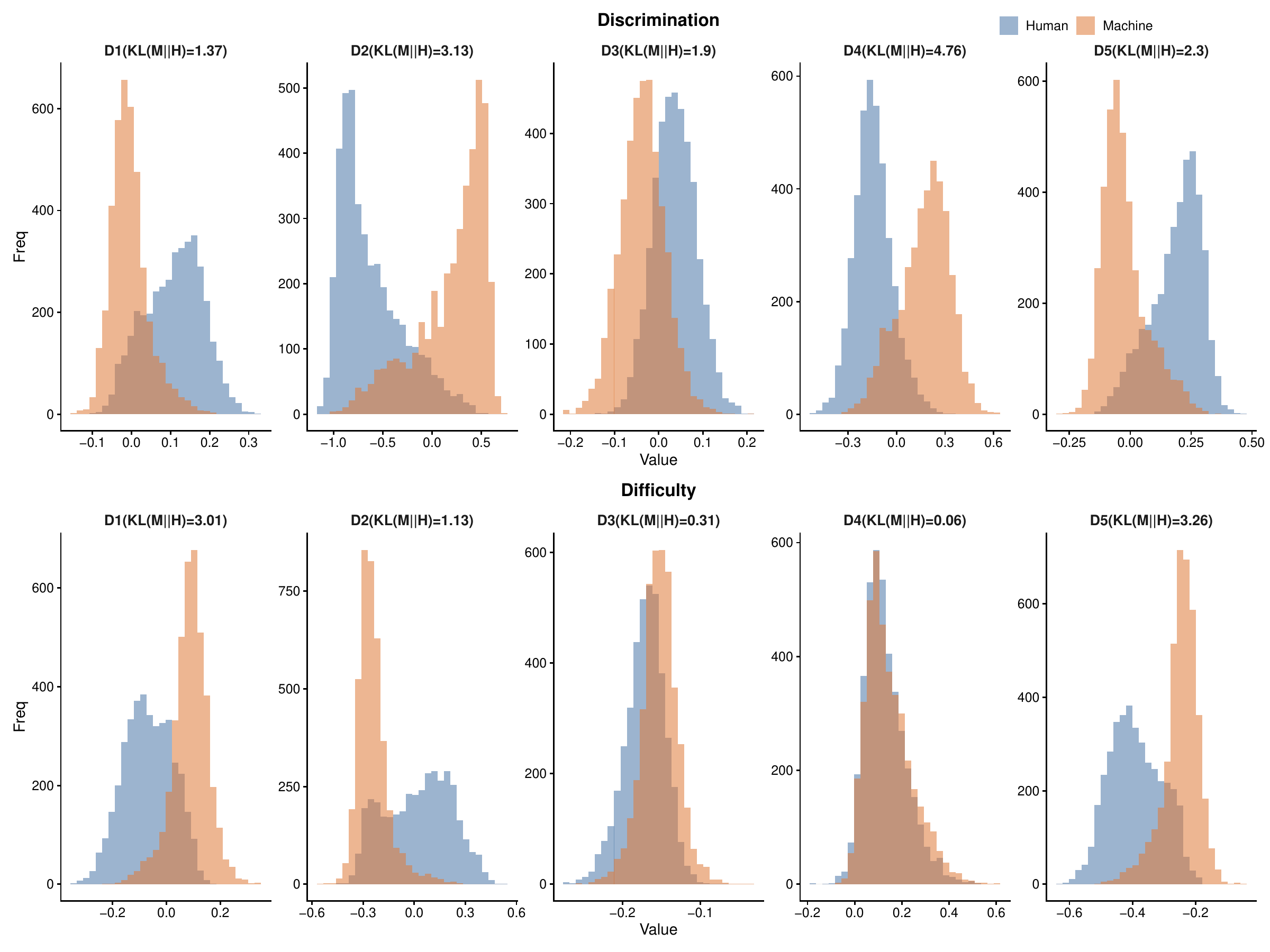}
    \caption{Marginal distributions of the estimated discrimination parameters $\alpha$ and difficulty-related location parameters $\beta$ for human-written and machine-generated texts in the XSum domain across the five latent dimensions. Each panel reports the corresponding domain-specific KL divergence $KL(M||H)$.}
    \label{fig:xsum}
\end{figure}

\begin{figure}[h]
     \centering
    \includegraphics[width=1\linewidth]{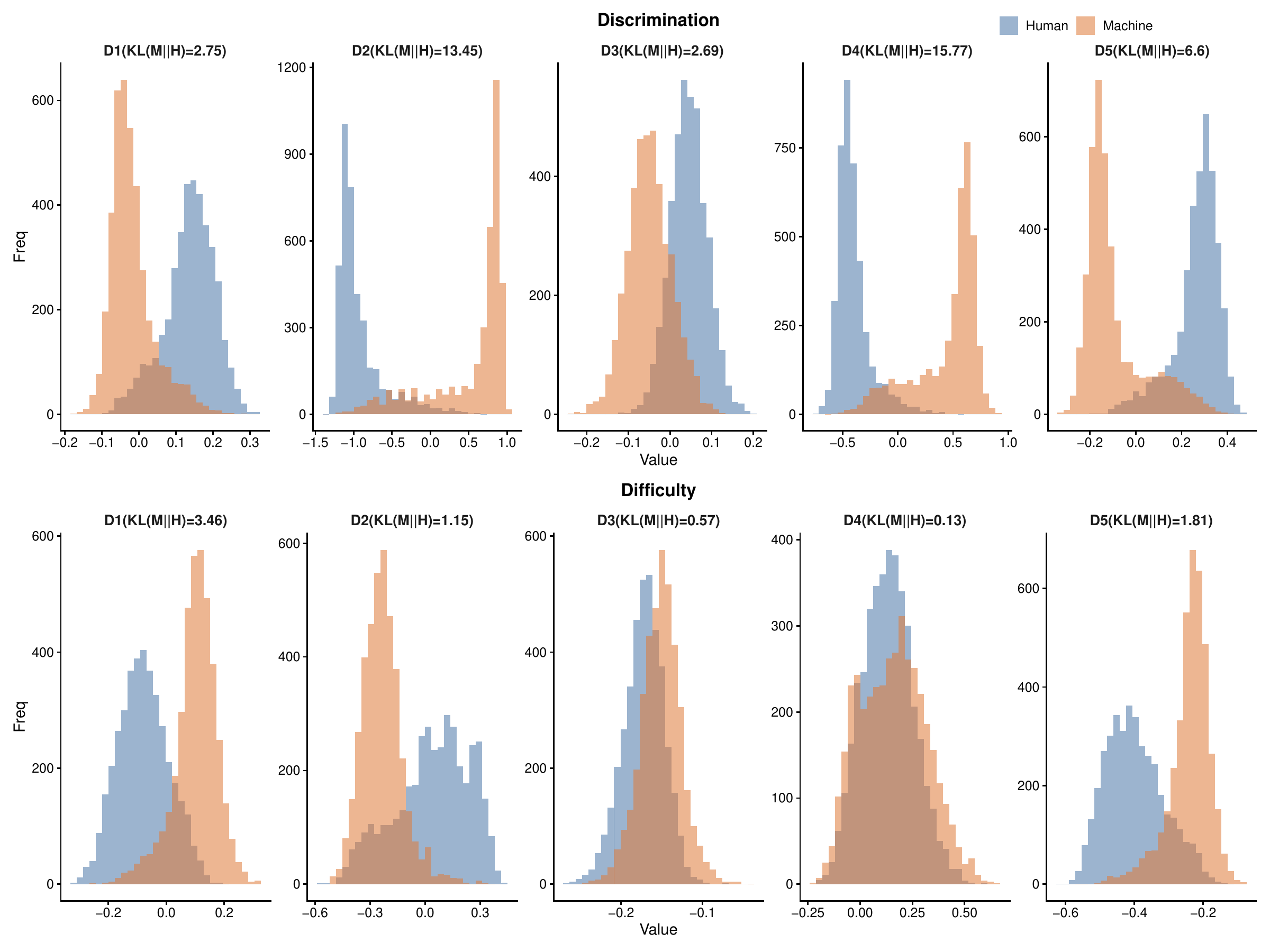}
    \caption{Marginal distributions of the estimated discrimination parameters $\alpha$ and difficulty-related location parameters $\beta$ for human-written and machine-generated texts in the Yelp Reviews domain across the five latent dimensions. Each panel reports the corresponding domain-specific KL divergence $KL(M||H)$.}
    \label{fig:yelp}
\end{figure}

The main text summarizes the domain-specific IRT structure using principal component visualizations and KL divergence values. Here, we provide additional dimension-wise distributions of the estimated discrimination parameters $\alpha$ and difficulty-related location parameters $\beta$ across individual domains. As shown in Figures \ref{fig:arxiv}-\ref{fig:yelp}, the degree of overlap between human-written and machine-generated texts varies across domains and latent dimensions. Some dimensions exhibit clear distributional shifts between the two groups, whereas others show substantial distributional overlap. These results indicate that the distinction between human-written and machine-generated texts is not uniformly distributed across the IRT dimensions. Instead, different domains exhibit distinct combinations of discrimination-related and difficulty-related distributional patterns.

These dimension-wise results provide additional support for the domain-specific patterns discussed in the main text. In particular, ArXiv exhibits greater distributional overlap across several dimensions, consistent with its weaker marginal separation and greater coverage asymmetry. By contrast, Writing Prompt, XSum, and Yelp Reviews exhibit more pronounced distributional differences in selected discrimination dimensions.

\subsection{Reverse KL divergence analysis}

To further examine the asymmetric distributional patterns between human-written and machine-generated texts, we compute the reverse-direction KL divergence $KL(H\|M)$ for each domain, where the human-written distribution is evaluated relative to the machine-generated distribution as the reference. This complements the main-text analysis based on $KL(M\|H)$, which measures how the machine-generated distribution diverges from the human-written reference.

As shown in Table \ref{tab:kl_rev}, the reverse-direction KL divergence values vary across domains and dimensions, indicating that the degree of distributional asymmetry depends on both domain and latent dimension. This pattern is particularly informative for ArXiv. In the main text, ArXiv exhibits only moderate $KL(M\|H)$ values, suggesting that machine-generated academic texts are not strongly separated from human-written academic texts in terms of their marginal IRT parameter distributions. However, the reverse direction, $KL(H\|M)$, yields relatively larger values in several dimensions, particularly $\beta_2$, $\beta_1$, $\beta_5$, and $\alpha_4$. This suggests that human-written ArXiv texts occupy regions of the IRT parameter space that are less well represented by machine-generated ArXiv texts.

\begin{table}
\caption{Domain-specific, length-stratified, and overall summary of $KL(H\|M)$ between human-written and machine-generated text distributions for the estimated IRT discrimination ($\alpha$) and difficulty-related location ($\beta$) parameters across five latent dimensions. The Overall row is computed from the pooled human-written and machine-generated distributions across all domains and is not the average of the domain-specific KL values.}
\centering\label{tab:kl_rev}
\begin{tabular}[h]{@{}l|l|rrrrr|rrrrr@{}}
\toprule
&&$\alpha_1$&$\alpha_2$&$\alpha_3$&$\alpha_4$&$\alpha_5$&$\beta_1$&$\beta_2$&$\beta_3$&$\beta_4$&$\beta_5$\\ 
\hline
\multirow{4}{*}{Domain}&ArXiv & 1.210 & 0.450 & 0.115 & 1.509 & 0.677 & 3.016 & 4.561 & 0.270 & 0.171 & 2.852\\
&Writing Prompt & 1.739 & 2.100 & 1.588 & 3.737 & 1.776 & 1.998 & 1.733 & 0.378 & 0.013 & 2.058\\
&XSum & 2.639 & 2.362 & 1.140 & 2.411 & 2.710 & 2.638 & 3.101 & 0.426 & 0.046 & 2.715\\
&Yelp & 2.162 & 3.255& 1.811 & 3.606 & 2.692 & 2.534 & 1.917 & 0.360 & 0.084 & 2.109\\
\hline
\multirow{3}{*}{Length}&Short &2.204& 2.756 & 1.465& 2.874& 2.890& 2.855& 4.173& 0.434& 0.022& 2.444\\
&Medium &1.544& 2.137 & 0.827& 2.184& 1.375& 1.752& 1.943& 0.297& 0.028& 3.057\\
 &Long &1.201& 1.768 & 1.200& 2.014& 1.551& 1.440& 1.387& 0.340& 0.017& 1.336\\
\hline
Overall &&1.012& 2.248 & 2.107& 0.812& 1.753& 1.469& 0.120& 2.164& 0.850& 1.763\\
\bottomrule
\end{tabular}
\end{table}

Therefore, the human-machine difference in ArXiv is better characterized by distributional asymmetry rather than complete separation between the two groups. In contrast, the other domains exhibit larger forward-direction KL divergences, $KL(M\|H)$, across several dimensions. The reverse KL divergence therefore provides complementary evidence of domain-specific distributional asymmetry without altering the overall interpretation of the main-text results.
\FloatBarrier
\section{Sensitivity analysis for compact benchmark construction}

\begin{figure}
    \centering
    \includegraphics[width=0.9\linewidth]{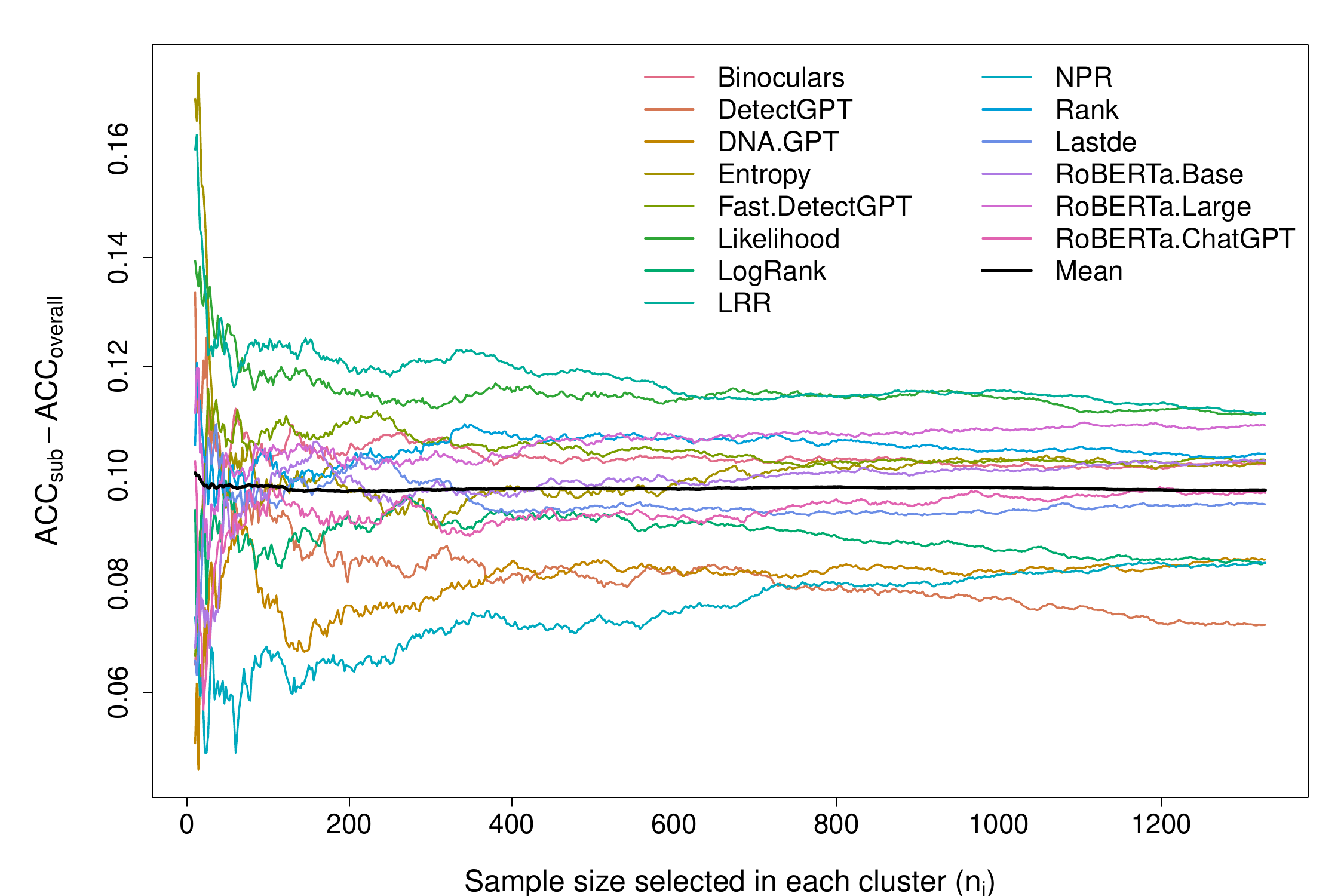}
    \caption{Detector-specific accuracy differences as the number of selected samples per cluster varies under $K=12$. Colored curves correspond to individual detectors, and the black curve represents the mean across detectors.}
    \label{fig:k12}
\end{figure}

Figure \ref{fig:k12} presents the detector-specific accuracy changes as the number of selected samples per cluster $n_i$ varies for $K=12$. The curves fluctuate more substantially when $n_i$ is small and gradually stabilize as more samples are selected, indicating that detector performance estimates become less sensitive to subset size once the selected subset reaches a sufficient size.

The stabilization of both the individual curves and the mean curve suggests that a relatively small representative subset can recover the main performance patterns observed on the full benchmark. In addition, the subset-based accuracies are slightly higher than the corresponding full-benchmark accuracies for several detectors. These modest differences may partly reflect the composition of the selected subset, particularly its emphasis on samples close to cluster centers in the IRT parameter space. Importantly, the purpose of the subset construction is not to improve detector performance, but to preserve the main evaluation patterns of the full benchmark with fewer samples. The differences across curves further indicate that detectors respond differently to changes in benchmark composition, highlighting heterogeneity in their sensitivity to sample selection.

\begin{figure}[h]
    \centering
    \includegraphics[width=1\linewidth]{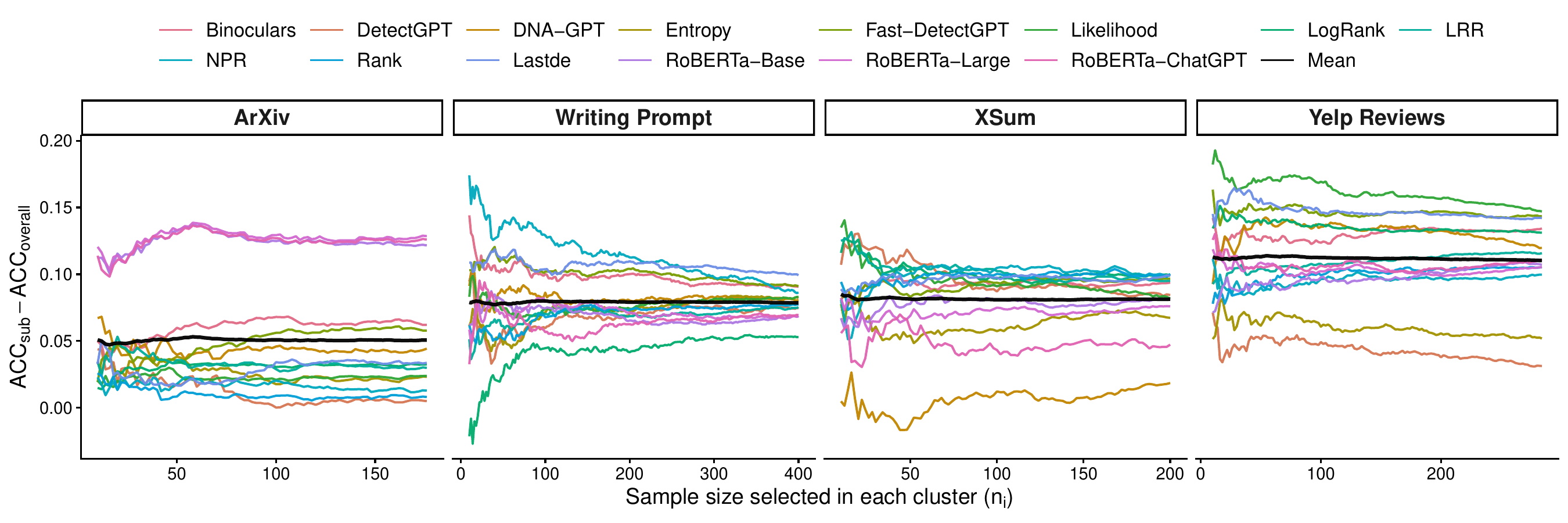}
    \caption{Domain-specific differences between subset-based and full-benchmark accuracies as the number of selected samples per cluster varies, with $K=16$ for ArXiv, $K=10$ for Writing Prompt, $K=15$ for XSum, and $K=14$ for Yelp Reviews. Colored curves represent individual detectors, and the black curve shows the mean across detectors.}
    \label{fig:domain_sel}
\end{figure}

Figure \ref{fig:domain_sel} extends this analysis to individual domains using domain-specific choices of $K$. Similar stabilization patterns are observed across domains, and the subset-based accuracies are slightly higher than the corresponding full-domain benchmark accuracies for most detectors. Although the magnitude of these differences varies across detectors and domains, the overall results suggest that relatively small representative subsets can preserve the main evaluation patterns of the full-domain benchmarks. The slightly higher accuracies observed for some detectors likely reflect differences in subset composition rather than systematic improvements in detector performance.

